%% file: main.tex
\documentclass[a4paper]{article}
\usepackage[T1]{fontenc}
\usepackage{setspace}
\usepackage{bbm}
\usepackage[pages=all, color=black, position={current page.south}, placement=bottom, scale=1, opacity=1, vshift=5mm]{background}
\SetBgContents{
 
}      % copyright

\usepackage[margin=1in]{geometry} % full-width
\usepackage{multirow}
\usepackage{booktabs}
\usepackage{makecell}
\usepackage{graphicx}
\usepackage{listings}
\usepackage{xcolor}
\usepackage{tcolorbox}
\tcbuselibrary{breakable}
\usepackage{tabularx}
\usepackage{seqsplit}
\usepackage{titlesec}
\usepackage{fvextra}
\usepackage{float}
\titleformat{\paragraph}[block]{\normalfont\bfseries}{}{0pt}{}
\titlespacing{\paragraph}{0pt}{1.2ex plus .1ex minus .1ex}{0.6em}

\lstdefinestyle{yamlstyle}{
  basicstyle=\ttfamily\small,
  breaklines=true,
  columns=fullflexible,
  backgroundcolor=\color{gray!5},
  frame=single,
  rulecolor=\color{gray!40}
}
\lstdefinestyle{pystyle}{
  basicstyle=\ttfamily\small,
  breaklines=true,
  columns=fullflexible,
  backgroundcolor=\color{gray!5},
  frame=single,
  rulecolor=\color{gray!40}
}
\lstdefinestyle{treestyle}{
  basicstyle=\ttfamily\small,
  breaklines=true,
  columns=fullflexible,
  backgroundcolor=\color{gray!5},
  frame=single,
  rulecolor=\color{gray!40}
}

\lstdefinestyle{bashstyle}{
  language=bash,
  backgroundcolor=\color{gray!8},
  basicstyle=\ttfamily\small,
  frame=single,
  rulecolor=\color{gray!50},
  breaklines=true,
  columns=fullflexible,
  keywordstyle=\color{blue!70!black}\bfseries,
  showstringspaces=false,
  xleftmargin=8pt,
  xrightmargin=8pt,
  framesep=6pt
}

\newcommand{\filename}[1]{\texttt{\textcolor{orange!100!black}{#1}}}

\newtcolorbox{promptbox}[1][]{
  breakable,
  colback=gray!5,
  colframe=gray!60,
  boxrule=0.5pt,
  arc=1mm,
  left=6pt, right=6pt, top=6pt, bottom=6pt,
  fonttitle=\bfseries,
  title=#1
}

\newtcolorbox{outputbox}[1][]{
  breakable,
  colback=teal!3,
  colframe=teal!45!gray!60,
  boxrule=0.5pt,
  arc=1mm,
  left=6pt, right=6pt, top=6pt, bottom=6pt,
  fonttitle=\bfseries,
  title=#1
}

\newtcolorbox{planbox}[1][]{
  breakable,
  colback=blue!3,
  colframe=blue!35!gray!55,
  boxrule=0.5pt,
  arc=1mm,
  left=6pt, right=6pt, top=6pt, bottom=6pt,
  fonttitle=\bfseries,
  title=#1
}
\usepackage{amsmath}
\usepackage{amsthm}
\usepackage{amssymb}

\usepackage[utf8]{inputenc}
\usepackage{hyperref}
\hypersetup{
	unicode,
}

\usepackage[sort&compress,numbers,square]{natbib}
\theoremstyle{plain}

\theoremstyle{definition}

\usepackage{lineno}
\usepackage{graphicx, color}
\graphicspath{{fig/}}

\usepackage{algorithm, algpseudocode} % use algorithm and algorithmicx for typesetting algorithms
\usepackage{mathrsfs} % for \mathscr command

\usepackage{lipsum}

\usepackage{color}

\title{Driving Epidemic Models with AI Agents: the Epydemix Agent Framework}

\author{
Nicolò Gozzi$^{1,2}$,
Ciro Cattuto$^{1}$\textsuperscript{*}, 
Alessandro Vespignani$^{2,1}$
}

\date{
$^1$ ISI Foundation, Turin, Italy\\
$^2$ Network Science Institute, Northeastern University, Boston, MA, USA\\
\textsuperscript{*}Corresponding author: Ciro Cattuto, ciro.cattuto@isi.it
}

\usepackage{url}

\begin{document}
	\maketitle

\begin{abstract}
Artificial Intelligence agents based on large language models provide convenient natural language interfaces to scientific software, but reliability is not automatic. Here we introduce the Epydemix Agent Framework, an additive layer over Epydemix, an open-source Python library for stochastic compartmental epidemic modeling. The framework extends the library with four capabilities to facilitate interaction with an AI agent: discovery of available models and parameters, preventive validation of a declarative scenario specification, execution through tested library code, and inspectability of results. These capabilities let an agent handle the entire modeling process, from the natural-language description of the scenario to quantitative results, figures, and interpretation of findings without writing custom code. Each step reads input files and saves results in a separate output bundle, making the process auditable and reproducible. First, we show the end-to-end workflow with a case study comparing vaccination strategies for a novel respiratory virus. Second, we assessed the framework across $50$ agent sessions and five modeling tasks by comparing the agent use of the framework against the direct use of the Python interface. The framework reduced turns, output tokens, and cost on most tasks, unless it trades resources for per-point reproducibility. 
\end{abstract}

\section{Introduction}
\label{sec:introduction}

Dynamic transmission models are widely used to simulate disease spread and to evaluate public health interventions~\cite{keeling2008modeling, metcalf2020mathematical, poletto2020applications, lofgren2014mathematical}. Realistic models simulate stochastic dynamics in age-structured populations, with contact matrices, and include interventions as changes in dynamics over time. Several infectious disease modeling tools exist as software libraries that expose programming interfaces, but their use requires programming expertise. Graphical user interfaces remove this requirement but constrain what a user can specify and simulate. The rise of Artificial Intelligence (AI) agents based on large language models (LLMs) offers a third mode of interaction, in which a researcher describes a task in natural language and an agent operates the software~\cite{Gottweis2026, Wang2023}. Current AI agents can translate natural-language task descriptions into sequences of operations performed via software tools~\cite{Lu2026}. As a result, they are now increasingly used to write code, perform analyses, and coordinate workflows. In epidemic modeling, AI agents allow a researcher to specify a scenario, configure a model, compare interventions, and interpret results without writing code for each analysis. Agents with such capabilities include Anthropic's Claude Code~\cite{claudecode2025}, OpenAI's Codex CLI~\cite{codexcli2025}, and open-source alternatives~\cite{opencode2025}.  

Epydemix~\cite{gozzi2025epydemix} is an open-source Python package for the development, simulation, and calibration of stochastic compartmental epidemic models. As with most scientific libraries, this package was not designed for text-mediated interaction. Instead, Epydemix was developed for scripts and computational notebooks, where configurations, model states, and results persist in memory as objects across function calls. An AI agent cannot reliably preserve such objects between tool calls. It has no way to discover library functionalities (e.g., available models and parameters) without reading source code or documentation. Similarly, an agent cannot extract summary measures such as attack rates from a binary output without writing custom code. Finally, sources of error may come from the agent itself. Indeed, it may specify parameter values that appear plausible but are invalid or reference functions and parameters that are not supported. These failure modes can have serious consequences in epidemic modeling, where outputs are used to compare interventions and mitigation strategies and can ultimately inform actions.

More generally, the reliable use of a modeling library from an AI agent requires four capabilities that a Python application programming interface (API) does not provide by default. The first one is \emph{discoverability}: programmatic access to available models and parameters, without requiring the agent to read source code or documentation. The second is \emph{declarative specification}: a self-contained description of a scenario that can be validated before execution. The third is \emph{reliable and reproducible execution}: running validated scenario configurations through calls to tested library code rather than through custom code the agent writes for the task. The last capability is \emph{inspectability}: compact textual summaries of simulation outputs, allowing the agent to obtain the quantities it needs without parsing raw outputs. These capabilities also support researchers' work. For instance, a declarative configuration can be inspected, edited, versioned, and shared. This also addresses a documented challenge of the field. A recent audit could reproduce only a small fraction of sampled COVID-era modeling studies~\cite{henderson2024reproducibility, pokutnaya2023implementation}, and several omissions were identified in reproducibility checklists~\cite{pokutnaya2023idmrc}.

In this paper, we introduce the Epydemix Agent Framework. The framework implements the four capabilities described above as an additive layer on top of the existing Epydemix Python API. This approach leaves the library and all existing notebooks and scripts unchanged. Its interface is stateless and file-based, so that each operation is a pure function of its input files. We demonstrate the end-to-end workflow with a case study comparing four intervention strategies for a novel respiratory virus, and we assess the framework across more than $50$ agent sessions.

\section{Materials and methods}
\label{sec:mem}
We use the term AI agent to refer to a system built around an LLM that operates external software through tool calling. A task is described to the agent through a natural language prompt~\cite{schick2023toolformer}. It then proceeds iteratively, selecting an operation, reading the returned result, and determining the next step until the task is complete~\cite{yao2022react}. The key property is that agents interact with software libraries entirely through text. The available functions, their arguments, and outputs must all be expressed as text and fit within the agents' context window.

\subsection{The Epydemix library}
\label{sec:epydemix-library}
Epydemix is an open-source Python package that supports the development and calibration of dynamic transmission models with any compartmental structure and dynamic public health interventions~\cite{gozzi2025epydemix}. To allow simulations with realistic demographic and mixing patterns, a companion repository including age-stratified distributions and social contact matrices is available~\cite{epydemix_data}. The package adopts a discrete-chain binomial process simulation engine and supports the full modeling pipeline, from model specification and simulation to parameter inference, forecasting, and scenario projection. The epidemic model is built as an object, with compartments and transitions added through method calls, and results returned as in-memory objects.

\subsection{The Epydemix Agent Framework}
\label{sec:mem:framework}
\begin{figure}[ht]
    \centering
\includegraphics[width=1\textwidth,keepaspectratio]{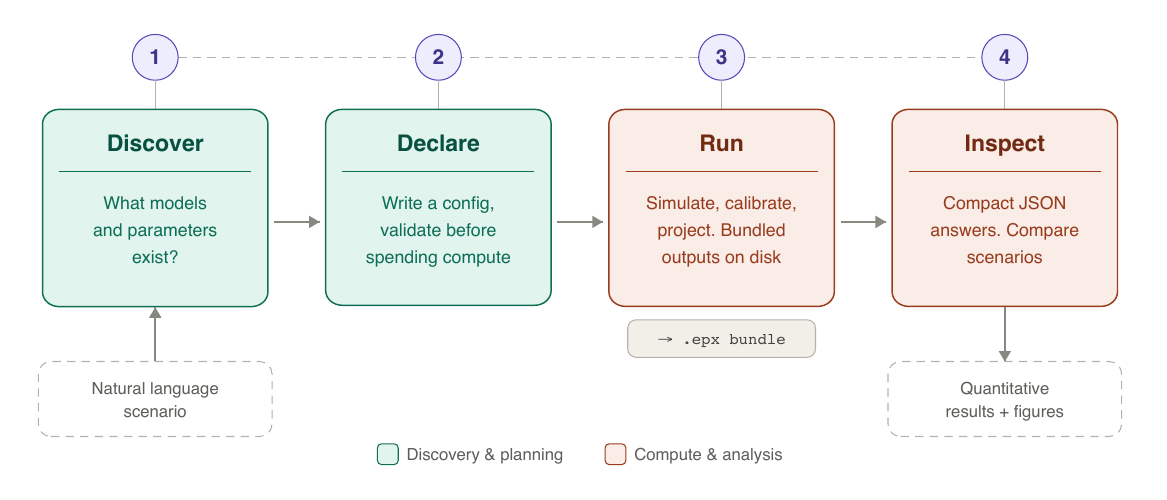}
    \caption{\textbf{Canonical four-phase workflow of the Epydemix Agent Framework.} Starting from a natural-language scenario description, an AI agent progresses through four sequential phases: (1) \textit{Discover}: enumerate available models and retrieve literature-sourced parameter defaults; (2) \textit{Declare}: encode the scenario as a YAML configuration file and validate it against the parameter registry before any computation; (3) \textit{Run}: execute simulation, calibration, or projection, writing results to a self-contained \texttt{.epx} output bundle; (4) \textit{Inspect}: query the bundle for compact, structured summaries and compare outcomes across scenarios. The framework is stateless: each phase operates exclusively on files, with no Python objects persisting across calls.}
    \label{fig:epydemix_agent_workflow}
\end{figure}

After being prompted with a natural-language scenario description, the agent framework proceeds through a four-phase workflow (Fig.~\ref{fig:epydemix_agent_workflow}). During the \emph{Discover} phase, the agent identifies the available models and retrieves default parameter values drawn from the literature for the disease of interest. During \emph{Declare}, the scenario is encoded in a configuration file (in YAML format) and validated against the parameter registry. Nothing is computed until validation passes. During \emph{Run}, the framework executes the simulation, calibration, or projection and writes the results to a self-contained output bundle. During \emph{Inspect}, the agent queries the output bundle for compact structured summaries and compares outcomes across scenarios. The current implementation uses the YAML format, but the design is not tied to any particular structured text format.

\begin{figure}[t]
  \centering
  \includegraphics[width=\columnwidth]{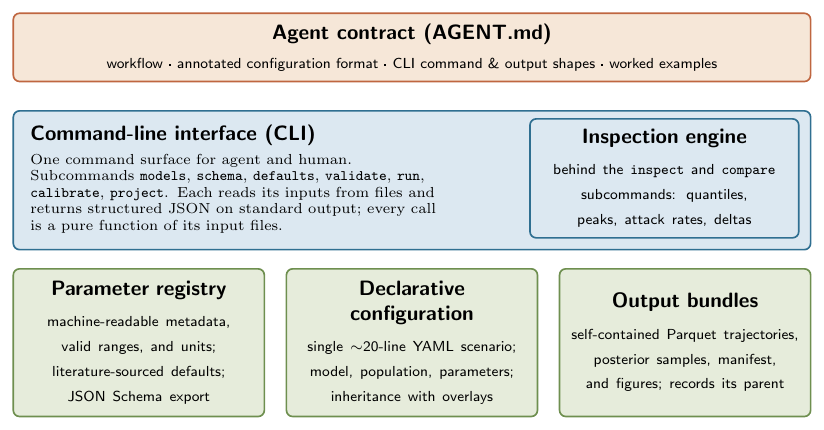}
  \caption{\textbf{The components of the Epydemix Agent Framework.} The agent contract (\texttt{AGENT.md}, top) is the document an agent reads first. The command-line interface (middle) is the single control surface for both an AI agent and a human researcher, and the inspection engine is part of it. The three data components (bottom) are the filesystem-backed artifacts that the framework reads and writes. The agent contract and tooling create a layer over the existing Epydemix Python API without modifying it.}
  \label{fig:epydemix_agent_workflow_components}
\end{figure}

Six components implement this workflow on top of the existing Epydemix Python API (Fig.~\ref{fig:epydemix_agent_workflow_components}). Three are on-disk artifacts the framework reads, writes, and inspects: the parameter registry, the declarative configuration files, and the output bundles. Two are the elements that operate on those artifacts, namely the command-line interface (CLI) and the inspection engine. The sixth, the agent contract, documents the framework in a form the agent can read and understand its use.

\textbf{Parameter registry.} Each model built from one of the predefined
base models (SIR, SEIR, SIS and SEIAR), together with any optional module layered on it (waning immunity, vaccination, clinical outcomes), carries a
machine-readable registry of its parameters. For every parameter the registry
records its name and a natural-language description of its role in the
dynamics, its kind (rate, probability, count, duration, proportion or
dimensionless), data type and expected shape (scalar, time-varying,
age-structured), its units, admissible range and default value,
whether it is required, and other parameters it depends on.  An agent can therefore discover what a model accepts without reading its source code.

\textbf{Declarative configuration.} A scenario configuration is a single, short YAML file that documents the model, the population (i.e., age distribution and contact matrices), the simulation time window, and the parameter values. The same file is the input to every downstream component. Configurations inherit through overlays, meaning that a scenario can import a base configuration and override only the selected fields. A comparison of four intervention strategies therefore translates into four short overlays on one single base file, which keeps the differences between them explicit. Configuration files can be automatically validated for correctness, so that, e.g., out-of-range values and inconsistent settings are detected before spending compute time. 

\textbf{Configuration-driven CLI.} The agent controls the workflow via a command-line interface (CLI) tool (\texttt{epydemix}). The CLI exposes each step of the workflow through dedicated sub-commands: \texttt{models}, \texttt{schema}, \texttt{defaults}, \texttt{validate}, \texttt{run}, \texttt{calibrate}, \texttt{project}, \texttt{inspect}, and \texttt{compare}. Each subcommand reads its inputs from files and writes standardized output into structured JSON. No state is carried between calls, so that every invocation is a pure function of its input files, and an agent or an automated pipeline can drive the workflow without a persistent Python session. We adopted statelessness because an agent cannot reliably hold Python objects across tool calls, while also letting a human script the same commands.

\textbf{Output bundles.} Every \texttt{run}, \texttt{calibrate}, and \texttt{project} operation returns its output in a standardized bundle enclosed within an \texttt{.epx} directory. The bundle contains simulation trajectories as Parquet files, posterior parameter samples for calibration runs, generated figures, and a \texttt{manifest.json} file recording the simulation metadata and source configuration. Each bundle also records its parent configuration and any upstream bundles it derives from. The provenance of a derived result can therefore be reconstructed from the filesystem alone, without the need for an external log.

\textbf{Inspection engine.} The inspection engine supplies the summarization logic behind the \texttt{inspect} and \texttt{compare} sub-commands. It converts the bundle's Parquet contents into compact JSON rather than returning raw arrays, reporting summary statistics of trajectories such as quantiles, peak timing and magnitude, attack rates, and pairwise differences between bundles. The agent thus obtains the quantities needed for the analysis of the simulations within its context window.

\textbf{Agent contract.} The document an agent reads first is the contract \texttt{AGENT.md}. The file describes the four-phase workflow, the configuration format with annotations for each field, the command-line interface with its CLI subcommands, and the exact structure of the JSON output. The contract also contains a set of worked-out examples. The first version of the contract was produced by having an agent inspecting the Epydemix source code, documentation, and the CLI built on top of it. The contract has then been refined iteratively by running the framework on an informal set of tasks, including the examples distributed with the library, and then using a second agent to review the execution logs and identify sources of friction. The revisions include a catalog of failure modes observed during development together with their corrections, and a translation guide mapping natural-language disease descriptions onto Epydemix compartments and transitions. The guide addresses the hardest step of the \emph{Declare} phase, where the model structure must be fixed before assigning any parameter values.

The declarative configuration format covers common compartmental modeling workflows but not the whole Epydemix API functionality. For example, a YAML configuration cannot express custom transition functions, non-standard age-group structures, or post hoc result transformations. Since the framework leaves the Epydemix API untouched, such functionalities remain available through direct API interaction, though they do not inherit the validation, provenance, and reproducibility of the structured interface. Indeed, the contract instructs the agent to use the CLI by default and to fall back on direct API calls only when an operation cannot be expressed declaratively.

The Epydemix agent framework is open source and available at \url{https://github.com/epistorm/epydemix}, under the same GPL-3.0 license as the core library.

\subsection{Experimental setup}
\label{sec:mem:setup}

\textbf{Software and installation.} All agent sessions used Claude Code from the terminal (version v2.1.198) with Claude Sonnet~5 as the underlying model, running under a Claude Pro subscription. We used version 1.3.1 of Epydemix and version 1 of the agent framework (commit hash 55d4e33). Sessions were run on July $1^{st}$, $2026$. The framework and its dependencies can be installed from a terminal via

\begin{lstlisting}[style=bashstyle]
git clone https://github.com/epistorm/epydemix.git
cd epydemix
pip install -e ".[agent]"
\end{lstlisting}

We recommend the installation inside a dedicated virtual environment, for example via \texttt{venv} or \texttt{conda}. This installs the \texttt{epydemix} command-line interface together with the agent contract. If the \texttt{epydemix} executable is not discoverable on \texttt{PATH} after installation, the CLI can equivalently be invoked as \texttt{python3 -m epydemix.cli.main} from the repository root. The framework requires that the agent can read local files and issue shell commands, and that it is invoked from the repository root so that \texttt{AGENT.md} is discoverable. A complete walkthrough for reproducing the case study, including installation, agent setup, and troubleshooting, is provided in the Supplementary Information (SI Guide~S4).

\textbf{Comparison conditions.} We run experiments considering two different conditions. In the \emph{framework} condition, the agent has access to the Epydemix CLI and to the agent contract, and is directed to follow the four-phase workflow described above. In the \emph{direct API} condition, the agent receives the same repository with all references to the CLI and the agent contract removed, and works through the Epydemix Python API directly with the library documentation and examples available. The direct API condition is therefore not in complete absence of tooling, but mimics the setup faced by a researcher using a well-documented library with a coding agent, without any additional layer mediating the interaction.

\textbf{Multi-task evaluation.} Our main assessment covers five epidemiological modeling tasks: an SEIRHD scenario comparison, an SIR calibration, a concatenated calibrate-then-project workflow, a school-closure sweep, and a measles coverage sweep. Each task was repeated five times in each of the two conditions, giving $50$ agent sessions in total. All sessions started from identical initial conditions, and a session could not access artifacts produced by another one. For each session, we extracted performance metrics from the agent's runtime system, namely wall-clock time, time spent in model requests, number of turns, number of output tokens, and cost. Task definitions and per-session values are given in the Supplementary Information (SI Additional Tasks~S5).

\textbf{Case study comparison.} We additionally ran the vaccination case study of Section~\ref{subsec:case-study} four times in each condition, giving eight further sessions. Each of these was split into two phases to separate agent exploration from execution. In the first phase, the agent operated in {\it plan} mode, reading the available context as well as the user's prompt and producing an inspectable plan. This phase was read-only, since tools that could modify state were unavailable. In this phase, no configuration files were written, and no simulations were run. In the second phase, the agent operated in {\it execute} mode and carried out the plan step by step. The split allows the cost and runtime of exploratory reading and reasoning to be attributed separately from those of the analysis itself.

\textbf{Metrics and statistics.} For each task and metric of the multi-task evaluation, we report the ratio of the median in the framework condition to the median in the direct API condition, so that values below one indicate lower resource use with the framework. Distribution-free confidence intervals for the ratio are obtained using the Wilcoxon rank-sum statistic. With five observations per condition, the attainable exact coverage is $96.8\%$, which is the level we report throughout. Intervals are computed per task and metric, without correction for multiple comparisons. For the case study comparison, we report the coefficient of variation across the four runs of each condition. We report the planning and execution phases separately. Costs are reported in US dollars as billed by the model provider at the time of the sessions. The cost should not be interpreted as a stable estimate and is given just to compare between conditions. 

\section{Results}
\label{sec:results}

\subsection{Case study: vaccination scenarios}
\label{subsec:case-study}

To illustrate the end-to-end workflow, we tasked an AI agent to compare vaccination strategies for a simulated epidemic. The study considered a novel respiratory virus spreading in Italy in a fully susceptible population. The disease assumptions include a basic reproduction number $\mathcal{R}_0 \approx 2.5$, a mean latent period of $3$ days, and a mean infectious period of $5$ days. We also assumed that, on average, $0.8\%$ of infections led to hospitalization, with a mean hospital stay of $10$ days, and that an available vaccine reduced susceptibility to infection by $85\%$. The simulations covered $16$ months and used $55,000$ beds as the national hospital-capacity threshold. The agent was instructed to include a no-vaccination baseline. Instructions on vaccination rates and the timing and magnitude of the non-pharmaceutical intervention in the combined strategy were left qualitative. This lack of detail is meant to test whether the agent can translate informal descriptions into internally consistent parameter values. The agent's input consists of the contract and the verbatim prompt below:

\begin{promptbox}[Prompt given verbatim to the agent]
\small
\textit{Read AGENT.md first, then carry out the following study end-to-end using the epydemix CLI. Work through the discover $\rightarrow$ declare $\rightarrow$ run $\rightarrow$ inspect workflow, use a shared base configuration with overlays for the scenarios, and validate every config before running it.}

\vspace{4pt}
\textbf{Scenario.} \textit{A novel respiratory virus is spreading in Italy in a fully susceptible population. Epidemiological assumptions: basic reproduction number R\textsubscript{0} $\approx$ 2.5, mean latent period $\approx$ 3 days, mean infectious period $\approx$ 5 days. About 0.8\% of infected individuals are hospitalized, with a mean hospital stay of $\approx$ 10 days. A vaccine is available that reduces transmission by $\approx$ 85\%. Italy can dedicate $\approx$ 55{,}000 hospital beds to this epidemic; sustained occupancy above that line means the health system is overwhelmed. Use a real Italian population with age-structured contact matrices (home, work, school, community) and simulate a 16-month horizon.}

\vspace{4pt}
\textit{\textbf{Important:} derive the \texttt{transmission\_rate} from the target R\textsubscript{0} using the population's contact structure --- do not use a preset transmission rate directly, since the contact matrix rescales it. Model hospitalization as a disease outcome and include a vaccination compartment.}

\vspace{4pt}
\textit{Compare four strategies:}
\begin{enumerate}
\item \textit{No vaccination --- the uncontrolled baseline.}
\item \textit{Slow rollout --- a low constant per-capita vaccination rate.}
\item \textit{Rapid rollout --- a high constant per-capita vaccination rate.}
\item \textit{Combined --- a moderate vaccination rate plus temporary non-pharmaceutical interventions (reduced school and workplace contacts for a few months during the wave).}
\end{enumerate}

\vspace{4pt}
\textit{For each strategy report: peak hospital occupancy, the peak date, the number of days occupancy exceeds the 55{,}000-bed capacity, and the total attack rate with deltas against the no-vaccination baseline. Then produce a single comparison figure showing hospital occupancy over time for all four strategies with the capacity line marked, and write a short interpretation: which strategies keep the system under capacity, and what the comparison implies for rollout speed versus combining measures.}
\end{promptbox}

We ran four instances of this study in the framework condition (Section~\ref{sec:mem:setup}). The agent made the same key modeling choices in all four instances. It selected an SEIR model with vaccination and hospitalization compartments and used the CLI's \texttt{schema} command to confirm that the model contained the compartments required by the scenarios. The transmission rate corresponding to the target $\mathcal{R}_0$ was derived considering the spectral radius of Italy's combined contact matrix ($\rho \approx 13.90$) instead of using a preset value. In every instance, the agent determined that the built-in attack-rate metric does not apply to a leaky-vaccine model and computed the attack rate instead from the sum of the susceptible-to-exposed and vaccinated-to-exposed transition flows, following the guidance in the contract. For the \textit{Combined} scenario, it ran the baseline first to locate the epidemic peak, then placed the temporary contact-reduction window around it.
As expected, the four runs diverged on quantities and choices the prompt left unspecified. For instance, each time the agent assigned different values for the per-capita vaccination rates, though their ordering across strategies was always preserved (\textit{Slow rollout} $<$ \textit{Combined} $<$ \textit{Rapid rollout}). The agent also used different numbers of stochastic realizations, simulation start date, and duration/magnitude of contact reduction in the \textit{Combined} scenario. 

\begin{figure}[h]
\centering
\includegraphics[width=0.85\textwidth]{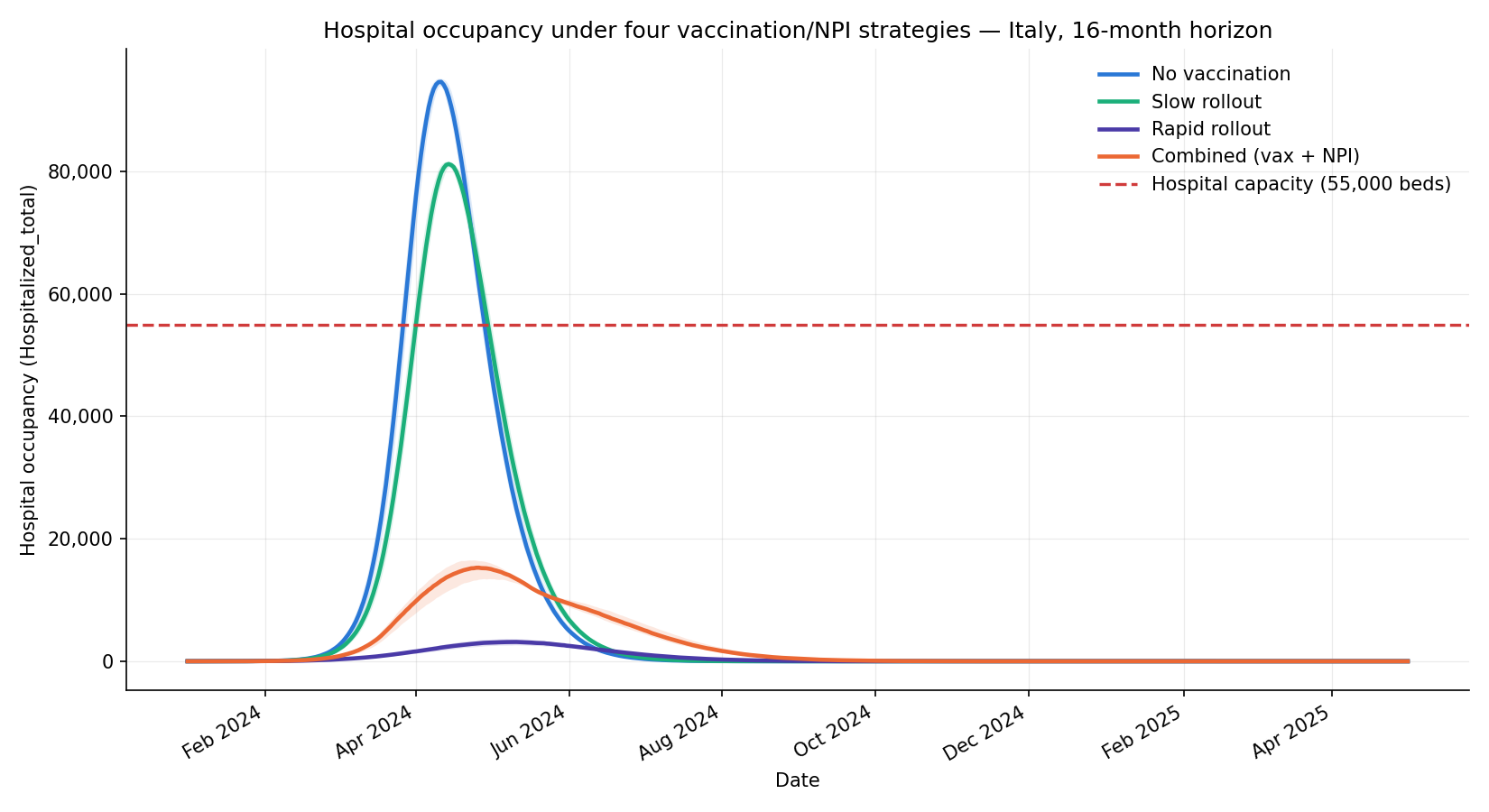}
\caption{\textbf{Hospital occupancy over the 16-month simulation horizon} under the four vaccination strategies, for the representative framework run reported in Section~\ref{subsec:case-study}, with the $55\text{,}000$-bed capacity threshold marked (dashed red line). Solid lines show the median across 100 stochastic simulations; shaded bands show the $90\%$ interval.}
\label{fig:vax-comparison}
\end{figure}

Figure~\ref{fig:vax-comparison} shows the hospital-occupancy trajectories, and the box below reproduces the agent's final report verbatim for one of the instances. The Supplementary Information (SI Plan~S1) includes the agent's plan, along with the layout, schema, and column-naming conventions of the underlying output bundle (SI Bundle~S2). In this run, the no-vaccination baseline exceeded hospital capacity for $33$ days and peaked at roughly $95\text{,}000$ occupied beds. \textit{Slow rollout} lowered the peak by about $13\text{,}500$ beds but still breached capacity for $29$ days. \textit{Rapid rollout} held peak occupancy below $3\text{,}500$ beds and the attack rate below $6\%$. The \textit{Combined} strategy, at half the vaccination rate of \textit{Rapid rollout}, also stayed below capacity, but its attack rate was $26.3\%$. That higher attack rate reflects transmission resuming after lifting the contact-reduction intervention, while vaccination continued at the lower rate.

\begin{outputbox}[Agent output]
\small
\textbf{Study complete}

\vspace{4pt}
\textbf{Setup:} SEIR + \texttt{vaccination} + \texttt{outcome: hospitalization} modules on Italy's real age-structured population (59.4M, 4 contact layers). \texttt{transmission\_rate = 0.0359742} derived from $\mathcal{R}_0=2.5$ via the contact matrix's spectral radius ($\rho\approx13.899$) rather than guessed. Shared \filename{base.yaml} + one overlay per scenario, all validated before running.
\vspace{4pt}
Bundles and configs live in \texttt{italy\_study/} (repo root).

\vspace{4pt}
\textbf{Results (deltas vs. no-vaccination baseline, Hospitalized\_total, 55,000-bed capacity)}
\small
\begin{tabularx}{\linewidth}{X rrrr}
\toprule
Metric & No vaccination & Slow rollout & Rapid rollout & Combined \\
 & (baseline) & (0.001/day) & (0.01/day) & (0.005/day + NPIs) \\
\midrule
Peak occupancy & 94,781 & 81,309 & 3,183 & 15,299 \\
 & & ($-$13,472) & ($-$91,598) & ($-$79,482) \\
\addlinespace
Peak date & 2024-04-10 & 2024-04-14 & 2024-05-08 & 2024-04-26 \\
 & & ($+$4d) & ($+$28d) & ($+$16d) \\
\addlinespace
Days over capacity & 33 & 29 ($-$4) & 0 ($-$33) & 0 ($-$33) \\
\addlinespace
Attack rate & 80.4\% & 72.7\% ($-$7.7 pp) & 5.6\% ($-$74.7 pp) & 26.3\% ($-$54.1 pp) \\
\bottomrule
\end{tabularx}
\vspace{0.2cm}
Combined scenario: rate 0.005/day + 70\% school / 30\% work contact reduction, 2024-03-20 to 2024-05-15.

\vspace{4pt}
\textbf{Figure}

\vspace{4pt}
\filename{italy\_study/no\_vax.epx/figures/hospital\_capacity\_comparison.png} (also registered in that bundle's manifest).

\vspace{4pt}
\textbf{Interpretation}

\vspace{4pt}
Only rapid rollout and the combined strategy keep the system under the 55,000-bed line --- slow rollout still overwhelms capacity for 29 days, barely better than doing nothing.
\vspace{4pt}
Rollout speed dominates: rapid vaccination alone (0.01/day, no NPIs) beats the combined strategy on every metric despite the combined scenario using a lower vaccination rate plus temporary contact reductions --- vaccinating fast enough to blunt the wave before it peaks is worth more than a moderate rate plus a few months of partial school/work closures.
\vspace{4pt}
The practical takeaway: if achievable, invest in rollout speed first; temporary NPIs are a useful hedge when rapid rollout capacity doesn't exist (moderate rate + NPIs still avoids capacity breach and cuts attack rate by more than half), but they are a substitute for speed, not a superior complement to it, given the specific rates and reduction assumptions used here.
\end{outputbox}

It is worth stressing that the figures in the agent's report carry more digits than the assumptions support. For instance, the derived transmission rate is reported to seven significant figures. For the sake of reproducing the agent results, we decided to report the verbatim output.

\subsection{Framework and direct API use compared}
\label{sec:results:comparison}

\subsubsection{The vaccination case study}
\label{sec:results:ablation}
\begin{figure}[h]
\centering
\includegraphics[width=0.95\textwidth]{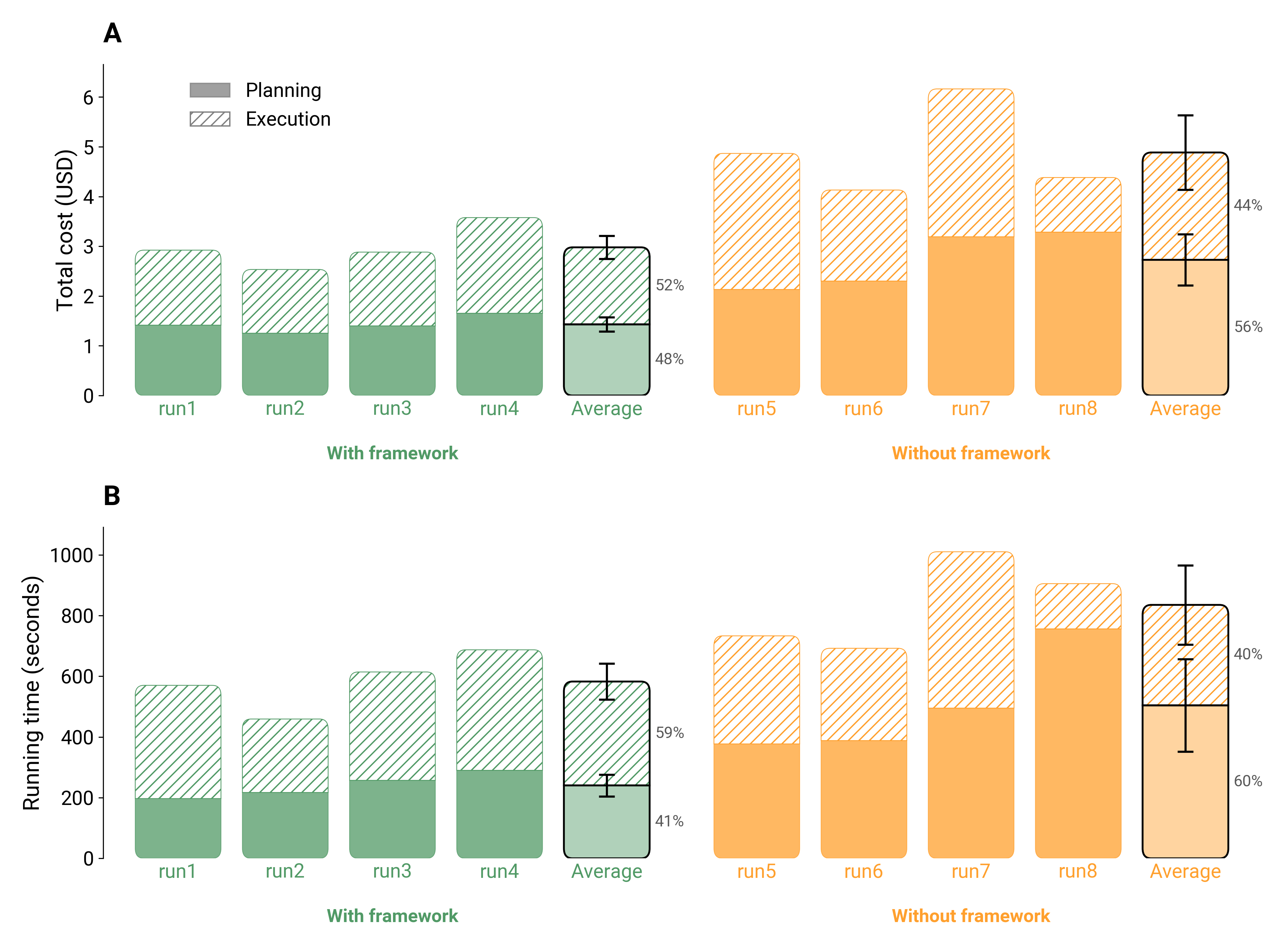}
\caption{\textbf{Compute-budget distribution across the eight case-study runs.} Top (A): total cost per run in USD, stacked by phase (solid: planning; hatched: execution), colored by condition (green: framework, run1--run4; orange: direct API, run5--run8). Bottom (B): wall-clock running time per run in seconds, stacked in the same way. The rightmost bar in each condition group shows the across-run average, with error bars giving $\pm 1$ standard deviation for each phase and the percentage of the total contributed by planning and execution annotated alongside.}
\label{fig:cost-turns}
\end{figure}

We first compare the two agent conditions, framework and direct API use, in the vaccination case study where the two-phase design separates exploration from execution, as presented in Section~\ref{subsec:case-study}. The two conditions reached similar qualitative epidemiological conclusions. In both conditions, the agent identified rollout speed as the main factor associated with outcomes, and used the same strategies to keep hospital occupancy below capacity. A detailed comparison of results and agent choices is given in the Supplementary Information (SI Comparison~S3). The mean total cost was \$2.98 in the framework condition and \$4.89 in the direct API condition, a $64\%$ increase. Mean total runtime was $582$~s with the framework and $835$~s without, an increase of $43\%$ (Fig.~\ref{fig:cost-turns}).
These totals are split between planning and execution. In the framework condition, the mean cost was divided $48\%$ to planning and $52\%$ to execution, on average. In the direct API condition, the shares were $56\%$ for planning and $44\%$ for execution. Runtime separated the conditions more sharply, with planning taking $41\%$ of total time under the framework and $60\%$ without it. The plan-mode transcripts show why. Agents in the direct API condition spent much of the planning phase reading the Epydemix source code for information that the \texttt{AGENT.md} contract provides to the framework condition. 

 The framework condition also shows lower Run-to-run variation (Table~\ref{tab:process-cv}). For cost, the coefficient of variation was $0.10$ during planning and $0.15$ during execution, against $0.19$ and $0.35$ in the direct API condition. For runtime, the corresponding values were $0.15$ and $0.17$ against $0.30$ and $0.40$. 

\begin{table}[h]
\centering
\begin{tabular}{llcc}
\toprule
Metric (CV = std/mean) & Phase & Framework & Direct API \\
\midrule
\multirow{2}{*}{Total cost}
 & Planning   & 0.10 & \textbf{0.19} \\
 & Execution  & 0.15 & \textbf{0.35} \\
\addlinespace
\multirow{2}{*}{Total duration}
 & Planning   & 0.15 & \textbf{0.30} \\
 & Execution  & 0.17 & \textbf{0.40} \\
\bottomrule
\end{tabular}
\caption{Coefficient of variation across the four case-study runs of each condition, for cost (USD) and running time (seconds), broken down by planning-phase and execution-phase values. Lower values indicate greater run-to-run consistency. The higher, less consistent value in each row is shown in bold.}
\label{tab:process-cv}
\end{table}

\subsubsection{Additional modeling tasks}
\label{sec:results:multitask}
To provide a more thorough analysis, we ran the comparison between agent conditions across five additional modeling tasks, with five runs per task in each condition. Across tasks, the framework generally reduced the computing resources the agent used (Table~\ref{tab:framework-ratios}). The \textit{Calibrate $\rightarrow$ Project} task shows the largest difference, with the framework condition completing the task in $0.53\times$ the wall-clock time, $0.45\times$ the output tokens, and $0.52\times$ the cost, with the confidence interval excluding parity on every metric. \textit{SIR calibration} and \textit{Measles coverage sweep} tasks show the same pattern. This suggests that a structured interface and an explicit contract remove much of the exploratory work. The agent spends fewer turns discovering how to set up and run a model and generates less text.

\begin{table}[h]
  \centering
  \small
  \begin{tabular}{lccccc}
    \toprule
    Task & Wall clock time (s) & Time in API (s) & \# Turns & \# Output tokens & Cost (\$) \\
    \midrule
    SEIRHD scenarios & \makecell{1.05\\[-1pt]{\scriptsize [0.85,\,4.78]}} & \makecell{0.98\\[-1pt]{\scriptsize [0.73,\,5.14]}} & \makecell{\textbf{0.77}\\[-1pt]{\scriptsize [0.60,\,0.96]}} & \makecell{\textbf{0.79}\\[-1pt]{\scriptsize [0.68,\,0.91]}} & \makecell{0.91\\[-1pt]{\scriptsize [0.81,\,1.04]}} \\
    \addlinespace[2pt]
    SIR calibration & \makecell{\textbf{0.73}\\[-1pt]{\scriptsize [0.59,\,0.95]}} & \makecell{\textbf{0.57}\\[-1pt]{\scriptsize [0.53,\,0.70]}} & \makecell{\textbf{0.62}\\[-1pt]{\scriptsize [0.55,\,0.72]}} & \makecell{\textbf{0.58}\\[-1pt]{\scriptsize [0.49,\,0.67]}} & \makecell{\textbf{0.71}\\[-1pt]{\scriptsize [0.65,\,0.95]}} \\
    \addlinespace[2pt]
    Calibrate $\rightarrow$ Project & \makecell{\textbf{0.53}\\[-1pt]{\scriptsize [0.45,\,0.57]}} & \makecell{\textbf{0.50}\\[-1pt]{\scriptsize [0.43,\,0.55]}} & \makecell{\textbf{0.63}\\[-1pt]{\scriptsize [0.53,\,0.70]}} & \makecell{\textbf{0.45}\\[-1pt]{\scriptsize [0.37,\,0.45]}} & \makecell{\textbf{0.52}\\[-1pt]{\scriptsize [0.49,\,0.73]}} \\
    \addlinespace[2pt]
    School-closure sweep & \makecell{\textbf{2.04}\\[-1pt]{\scriptsize [1.24,\,3.27]}} & \makecell{\textbf{1.41}\\[-1pt]{\scriptsize [1.25,\,2.24]}} & \makecell{1.38\\[-1pt]{\scriptsize [1.00,\,2.21]}} & \makecell{\textbf{1.32}\\[-1pt]{\scriptsize [1.18,\,2.23]}} & \makecell{\textbf{1.81}\\[-1pt]{\scriptsize [1.34,\,3.34]}} \\
    \addlinespace[2pt]
    Measles coverage sweep & \makecell{0.81\\[-1pt]{\scriptsize [0.42,\,1.12]}} & \makecell{\textbf{0.83}\\[-1pt]{\scriptsize [0.62,\,0.93]}} & \makecell{\textbf{0.80}\\[-1pt]{\scriptsize [0.68,\,0.89]}} & \makecell{\textbf{0.71}\\[-1pt]{\scriptsize [0.62,\,0.93]}} & \makecell{0.88\\[-1pt]{\scriptsize [0.67,\,1.05]}} \\
    \addlinespace[2pt]
    \bottomrule
  \end{tabular}
  \caption{Framework performance relative to direct use of the Epydemix Python API, by task. Each cell reports the ratio of medians (framework\,/\,direct API) over $5$ runs per condition, together with an exact $96.8\%$ distribution-free confidence interval for the ratio (Wilcoxon rank-sum). Values below $1$ indicate lower resource use in the framework condition. Values in bold indicate ratios whose confidence interval excludes $1$.}
  \label{tab:framework-ratios}
\end{table}

The comparison between the two operating conditions of the agent is, however, strongly task-dependent. Interestingly, the \textit{School-closure sweep} task reverses the pattern. For this task, the framework condition requires $1.41\times$ the API time, $1.32\times$ the output tokens, and $1.81\times$ the cost relative to the direct API condition. The origin of this reversal can be traced back to the artifacts the agent produced. The task sweeps over $18$ intervention start dates. The framework condition considers each sweep as an independent run with its own configuration file and a self-describing bundle. In the direct API condition, the same $18$ sweeps run as a single Python loop that writes multiple output files. This implies that the higher use of resources in the framework condition is not an efficiency loss. On the contrary, resources are spent to have every step of the sweep recorded as an independent and auditable artifact, storing the configuration that produced it, its outputs, a provenance record, and the random seed used for its generation. Re-executing a bundle's stored configuration reproduces that point's realizations, so the sweep is reproducible point by point and not only in aggregate. The direct API sweep is not unreproducible per se, but data preservation varies from run to run. In our testing across runs, it produced different result schemas each time and retained raw per-point trajectories in only three cases out of five. In summary, in this case the framework's use of resources should be thought of as a trade-off for reproducibility and not a performance regression.

\section{Discussion}
\label{sec:discussion}
The Epydemix Agent Framework shows how an infectious disease modeling library can be reliably operated by an AI agent through an additive interface. Across modeling tasks, agents working within the framework generally used fewer turns, fewer output tokens, and less compute than agents working directly with the Python API. In the vaccination case study, both working conditions reached the same qualitative conclusions, but the framework condition was cheaper, faster, and more consistent from run to run. 

The framework is designed in a way that allows validation against the parameter registry before the simulation begins. The specification stage catches invalid configurations, so users can correct them before spending compute. This mechanism explains the shift from planning toward execution seen in the case-study comparison (Section~\ref{sec:results:ablation}). The validated, self-contained configuration removes the need to verify inputs through exploratory reading. 
The \textit{School-closure sweep} task shows that the framework condition could cost more than direct API use, because it ran $18$ sweep points as independent bundles. The extra cost pays off in the output's properties rather than in a better result. Each point of the sweep survives with its configuration, its seed, and its provenance, so the sweep is reproducible point by point. In the direct API condition, that property held in just three out of five runs, and the output schema differed in all five. This does not necessarily imply that the agent committed an error. However, structured interfaces convert such properties (e.g., reproducibility) from a matter of the agent's choices into a matter of consistency of the interface. Whether the cost is worth paying depends on the specific context.

Here we implemented library control using a text-based command-line interface (CLI) rather than, for example, using a Model Context Protocol (MCP) server~\cite{anthropic2024mcp}. MCP standardizes tool discovery and invocation through a client-server connection managed by the agent host, and the state of the server needs to be managed. A CLI, instead, affords a stateless, file-based design: inputs and outputs can be inspected and versioned, and the same commands run equally well for a human or an agent, with no separate server process. If needed, nothing prevents the CLI from being exposed via an MCP server later, without changing the workflow beneath it.

Our study has limitations stemming from the analysis of a single agentic coding tool and the limited set of tasks used to test the framework. Moving from the demonstrations provided here to a systematic evaluation of AI agent-driven modelling approaches will require community-defined benchmarks. In particular, the definition of a representative set of infectious disease modeling tasks with specific measures of correctness and reproducibility. Such a benchmark would turn potential failure modes documented in the agent contract into measurable outcomes, allow different models and agent tooling to be compared on the same tasks, and show which parts of this framework can be generalized beyond Epydemix. Whether open-weight models perform comparably to closed ones~\cite{claudecode2025, codexcli2025, opencode2025} should be tested on a community-defined benchmark. 

As AI agents become a common way to operate scientific software, interfaces designed for them, rather than for interactive programming sessions, can play a key role in making agent-driven epidemic modeling efficient, auditable, and reproducible.

\section*{Acknowledgements}
NG, CC and AV acknowledge partial support from the Lagrange Project of the ISI Foundation, funded by CRT Foundation. AV acknowledges support from the cooperative agreement CDC-RFA-FT-23-0069 from the CDC’s Center for Forecasting and Outbreak Analytics. The findings and conclusions in this study are those of the authors and do not necessarily represent the official position of the funding agencies, the CDC, or the U.S. Department of Health and Human Services of the United States.

\section*{Declaration of AI and AI-assisted technologies use}
During the preparation of this work the author(s) used agent sessions in Claude Code from the terminal (version v2.1.198) with Claude Sonnet~5 as the underlying model, running under a Claude Pro subscription. We report verbatim prompt and output of one experiment. AI-assisted technologies were utilized for proofreading, text editing, and structural formatting. Specifically, these tools were employed to enhance grammatical clarity, improve readability, and align the text with the journal's style guidelines. All AI-generated suggestions were thoroughly reviewed and verified by the authors, who maintain full responsibility for the final content.

\section*{Authors contributions}

\section*{Competing interests}
Authors declare no competing interests.

\section*{Data availability}

The Epydemix Agent Framework is open-source software, released under the GPL-3.0 license, and is available at \url{https://github.com/epistorm/epydemix}. The agent contract (\texttt{AGENT.md}), CLI source code, and documentation described in this paper are included in that repository. The full prompt and agent output for the case study reported in Section~\ref{sec:results} are reproduced verbatim in the main text; the agent's plan and a step-by-step guide to reproduce the case study from scratch are reported verbatim in the Supplementary Information (SI Plan~S1, SI Guide~S4).

\bibliography{refs}

% ================= SUPPLEMENTARY INFORMATION =================
\clearpage
\setcounter{table}{0}
\setcounter{figure}{0}
\setcounter{lstlisting}{0}
\renewcommand{\thetable}{S\arabic{table}}
\renewcommand{\thefigure}{S\arabic{figure}}
\renewcommand{\thelstlisting}{S\arabic{lstlisting}}
\begin{center}
{\LARGE Supplementary Information}
\end{center}

\section*{SI Plan S1: Agent plan (Claude Code, plan mode)}
\label{si:plan-s1}

The following is the plan produced verbatim by the agent when invoked in plan mode, prior to any execution.

\begin{planbox}[Agent plan]
\small

\subsubsection*{Study: Novel Respiratory Virus in Italy --- Vaccination Strategies}

\paragraph{Context.}
The user wants an end-to-end \texttt{epydemix} CLI study of a novel respiratory virus spreading in a fully susceptible Italian population, comparing four vaccination/NPI strategies (no vaccination, slow rollout, rapid rollout, combined rollout+NPIs) on hospital-capacity and attack-rate outcomes. The workflow must follow discover $\rightarrow$ declare $\rightarrow$ run $\rightarrow$ inspect, use a shared base config with scenario overlays, validate every config before running, and derive \texttt{\seqsplit{transmission\_rate}} from the target $\mathcal{R}_0$ via the population's contact-matrix spectral radius (not a preset value). Deliverable: a metrics table (deltas vs. no-vaccination baseline) and a single comparison figure of hospital occupancy with the capacity line, plus a short written interpretation.

Discovery already performed (read-only, in this session):
\begin{itemize}
  \item \texttt{epydemix models} confirms SEIR is available; CLI binary lives at \texttt{\seqsplit{/opt/anaconda3/bin/epydemix}} (not on default PATH --- must prepend \texttt{\seqsplit{/opt/anaconda3/bin}} or call by full path in every command).
  \item \texttt{epydemix schema SEIR -{}-vaccination -{}-outcome hospitalization} confirms the modular composition gives exactly the compartments/parameters needed: \texttt{Susceptible, Exposed, Infected, Recovered, Vaccinated, Hospitalized}, params \texttt{\seqsplit{transmission\_rate}, \seqsplit{incubation\_rate}, \seqsplit{recovery\_rate}, \seqsplit{vaccination\_rate}, \seqsplit{vaccine\_efficacy}, \seqsplit{hospitalization\_rate}, \seqsplit{hospitalization\_recovery\_rate}}. \texttt{\seqsplit{vaccination\_rate}} is a constant per-capita S$\rightarrow$V rate (not a scheduled dose campaign) --- exactly matches ``low/high constant per-capita vaccination rate'' in the prompt. \texttt{\seqsplit{vaccine\_efficacy}} reduces breakthrough transmission (V$\rightarrow$Exposed since backbone has E), matching ``vaccine reduces transmission by $\sim$85\%''.
  \item A throwaway config with this exact module combination + \texttt{population: Italy} validated successfully (\texttt{valid: true}), confirming compartment naming (\texttt{\seqsplit{Susceptible/Exposed/Infected/Recovered/Vaccinated/Hospitalized}}) for \texttt{\seqsplit{initial\_conditions}}.
  \item Computed Italy's combined contact-matrix spectral radius directly (per the recipe in \filename{AGENT.md} ``Parameter Discovery''): $\rho \approx 13.8988$, population total $59{,}435{,}140$ across groups \texttt{['0-4','5-19','20-49','50-64','65+']}. $\texttt{\seqsplit{transmission\_rate}} = \texttt{R0\_target} \times \texttt{\seqsplit{recovery\_rate}} / \rho = 2.5 \times 0.2 / 13.8988 \approx 0.035974$.
\end{itemize}

\paragraph{Derived / assumed parameters.}

\small
\begin{tabularx}{\linewidth}{l l X}
\toprule
Quantity & Value & Source \\
\midrule
\texttt{\seqsplit{transmission\_rate}} & \texttt{0.035974} & derived from R$_0$=2.5, recovery\_rate=0.2, Italy $\rho\approx13.8988$ (computed above) \\
\texttt{\seqsplit{incubation\_rate}} & \texttt{0.333333} & 1 / mean latent period (3 days) \\
\texttt{\seqsplit{recovery\_rate}} & \texttt{0.2} & 1 / mean infectious period (5 days) \\
\texttt{\seqsplit{hospitalization\_rate}} & \texttt{0.008} & 0.8\% of infected hospitalized \\
\texttt{\seqsplit{hospitalization\_recovery\_rate}} & \texttt{0.1} & 1 / mean hospital stay (10 days) \\
\texttt{\seqsplit{vaccine\_efficacy}} & \texttt{0.85} & 85\% transmission reduction \\
Capacity line & \texttt{55000} beds & given \\
\texttt{\seqsplit{vaccination\_rate}} (no vaccination) & \texttt{0.0} & baseline \\
\texttt{\seqsplit{vaccination\_rate}} (slow rollout) & \texttt{0.001} & ``low'' constant per-capita rate ($\sim$1/1000 day) \\
\texttt{\seqsplit{vaccination\_rate}} (rapid rollout) & \texttt{0.01} & ``high'' constant per-capita rate ($\sim$1/100 day) \\
\texttt{\seqsplit{vaccination\_rate}} (combined) & \texttt{0.005} & ``moderate'', between slow and rapid \\
NPI reduction (combined only) & school 70\%, work 30\% & temporary, 8 weeks during the wave \\
\bottomrule
\end{tabularx}
\vspace{0.1cm}

Simulation horizon: \texttt{\seqsplit{start\_date: 2024-01-01}}, \texttt{\seqsplit{end\_date: 2025-05-01}} (16 months), \texttt{dt: 0.5}. \texttt{\seqsplit{n\_simulations: 100}} per scenario (matches \filename{AGENT.md} example conventions; compartmental model at 5 age groups is cheap regardless of the 59M population size).

Initial conditions (fraction mode, mirrors the Italy-population convention already used in \filename{AGENT.md}'s SEIAR example --- $\sim$100 people seeded in each of E and I):

\begin{lstlisting}[style=yamlstyle]
Susceptible: 0.9999966350
Exposed:     0.0000016825
Infected:    0.0000016825
Recovered:   0.0
Vaccinated:  0.0
Hospitalized: 0.0
\end{lstlisting}

\paragraph{Directory layout.}
Create under the repo (not scratch, since this is a real deliverable the user will want to keep and revisit):

\begin{lstlisting}[style=treestyle]
epydemix/studies/italy_respiratory_virus/
  configs/
    base.yaml
    scenario_no_vaccination.yaml
    scenario_slow_rollout.yaml
    scenario_rapid_rollout.yaml
    scenario_combined.yaml
  bundles/
    no_vaccination.epx/
    slow_rollout.epx/
    rapid_rollout.epx/
    combined.epx/
  analyze.py            # attack-rate + figure generation script
\end{lstlisting}

\paragraph{Steps.}

\textbf{1. Write \filename{\seqsplit{configs/base.yaml}}} --- shared model/population/simulation/initial\_conditions:

\begin{lstlisting}[style=yamlstyle]
model:
  type: SEIR
  vaccination: true
  outcome: hospitalization
parameters:
  transmission_rate: 0.035974
  incubation_rate: 0.333333
  recovery_rate: 0.2
  vaccination_rate: 0.0
  vaccine_efficacy: 0.85
  hospitalization_rate: 0.008
  hospitalization_recovery_rate: 0.1
population:
  name: "Italy"
  contact_layers: ["home", "work", "school", "community"]
simulation:
  start_date: "2024-01-01"
  end_date: "2025-05-01"
  dt: 0.5
  n_simulations: 100
initial_conditions:
  Susceptible: 0.9999966350
  Exposed: 0.0000016825
  Infected: 0.0000016825
  Recovered: 0.0
  Vaccinated: 0.0
  Hospitalized: 0.0
\end{lstlisting}

\textbf{2. Write the three non-dependent overlays} (each: \texttt{base: base.yaml} + a \texttt{parameters:} key-merge overriding only \texttt{\seqsplit{vaccination\_rate}}):
\begin{itemize}
  \item \filename{\seqsplit{scenario\_no\_vaccination.yaml}} $\rightarrow$ \texttt{\seqsplit{vaccination\_rate: 0.0}} (explicit, self-documenting)
  \item \filename{\seqsplit{scenario\_slow\_rollout.yaml}} $\rightarrow$ \texttt{\seqsplit{vaccination\_rate: 0.001}}
  \item \filename{\seqsplit{scenario\_rapid\_rollout.yaml}} $\rightarrow$ \texttt{\seqsplit{vaccination\_rate: 0.01}}
\end{itemize}

\textbf{3. Validate and run these three first} (in dependency order --- \texttt{\seqsplit{no\_vaccination}} must run before we can time the combined scenario's NPI window):

\begin{lstlisting}[style=bashstyle]
epydemix validate configs/scenario_no_vaccination.yaml
epydemix run configs/scenario_no_vaccination.yaml -o bundles/no_vaccination.epx
epydemix validate configs/scenario_slow_rollout.yaml
epydemix run configs/scenario_slow_rollout.yaml -o bundles/slow_rollout.epx
epydemix validate configs/scenario_rapid_rollout.yaml
epydemix run configs/scenario_rapid_rollout.yaml -o bundles/rapid_rollout.epx
\end{lstlisting}

\textbf{4. Determine the NPI window for the combined scenario:}

\begin{lstlisting}[style=bashstyle]
epydemix inspect bundles/no_vaccination.epx peak -v Hospitalized_total
\end{lstlisting}

Set the intervention window to roughly 4 weeks before through 4 weeks after the median peak date ($\approx$8 weeks total, ``a few months during the wave''), clipped to stay within the simulation horizon.

\textbf{5. Write \filename{\seqsplit{scenario\_combined.yaml}}:}

\begin{lstlisting}[style=yamlstyle]
base: base.yaml
parameters:
  vaccination_rate: 0.005
interventions:
  - layer: school
    start_date: "<peak - 4w>"
    end_date: "<peak + 4w>"
    reduction: 0.7
  - layer: work
    start_date: "<peak - 4w>"
    end_date: "<peak + 4w>"
    reduction: 0.3
\end{lstlisting}

Validate and run:

\begin{lstlisting}[style=bashstyle]
epydemix validate configs/scenario_combined.yaml
epydemix run configs/scenario_combined.yaml -o bundles/combined.epx
\end{lstlisting}

\textbf{6. Inspect / compare --- built-in metrics} (peak occupancy, peak date, days over capacity), with deltas vs. the no-vaccination baseline:

\begin{lstlisting}[style=bashstyle]
epydemix compare bundles/no_vaccination.epx bundles/slow_rollout.epx \
  bundles/rapid_rollout.epx bundles/combined.epx \
  -n NoVaccination,SlowRollout,RapidRollout,Combined \
\end{lstlisting}

\textbf{7. Attack rate --- custom Python} (per \filename{AGENT.md}: \texttt{\seqsplit{attack\_rate}} built-in metric is wrong for vaccination models; this is a leaky vaccine with a V$\rightarrow$Exposed breakthrough transition, so both \texttt{\seqsplit{S\_to\_E\_total}} and \texttt{\seqsplit{V\_to\_E\_total}} flows must be summed):

\begin{lstlisting}[style=pystyle]
import pandas as pd
N = 59_435_140  # Italy total population, from population.total_population
for name, bundle in scenarios.items():
    df = pd.read_parquet(f"{bundle}/transitions.parquet",
                          columns=["sim_id", "S_to_E_total", "V_to_E_total"])
    attack = df.groupby("sim_id")[["S_to_E_total", "V_to_E_total"]].sum().sum(axis=1) / N * 100
    # report median + 90% CI, and delta vs. no_vaccination median
\end{lstlisting}

Assemble a single summary table: peak occupancy, peak date, days over 55,000, attack rate --- each as a delta vs. no-vaccination baseline --- for the three vaccinated scenarios (and absolute values for the baseline itself).

\textbf{8. Comparison figure} (\filename{\seqsplit{analyze.py}}, adapting \filename{AGENT\_EXAMPLES.md} Recipe 2 + Recipe 3): overlay median \texttt{\seqsplit{Hospitalized\_total}} (with a light 90\% CI band) for all four scenarios, \texttt{axhline} at 55,000 with a ``Capacity (55,000 beds)'' label. Save to \filename{\seqsplit{bundles/no\_vaccination.epx/figures/hospital\_capacity\_comparison.png}}, then register with \texttt{\seqsplit{add\_figure\_to\_manifest(..., description=..., variables=["Hospitalized\_total"])}}. Report the absolute figure path to the user.

\textbf{9. Interpretation} --- short written summary (in chat, not a new file) covering: which strategies keep median occupancy under 55,000 beds and for how long any exceed it; how peak occupancy and attack rate scale with rollout speed (slow vs. rapid, holding NPIs fixed); and what the combined scenario's delta implies about moderate vaccination + temporary NPIs vs. vaccination speed alone (i.e., does combining beat rapid-alone, and by how much, on peak occupancy / days-over-capacity / attack rate).

\paragraph{Verification.}
\begin{itemize}
  \item Every config passes \texttt{epydemix validate} with \texttt{"valid": true} before it is run.
  \item After each \texttt{run}, confirm the bundle's \texttt{\seqsplit{manifest.json}} lists the expected compartment columns (\texttt{\seqsplit{Susceptible\_total, Exposed\_total, Infected\_total, Recovered\_total, Vaccinated\_total, Hospitalized\_total}}) via \texttt{epydemix inspect <bundle> manifest}.
  \item Sanity-check the no-vaccination baseline's dynamics look like a single wave with total attack rate broadly consistent with $1 - 1/R_0$-scale expectations for R$_0$=2.5 ($\sim$85--90\% of an unmitigated fully-susceptible population, tempered by the age-structured contact matrix) rather than something degenerate (e.g., near-0\% or near-100\% instantly), per \filename{AGENT.md}'s warning to sanity-check dynamics after deriving \texttt{\seqsplit{transmission\_rate}}.
  \item Confirm \texttt{epydemix compare} deltas match hand-checked arithmetic for at least one metric.
  \item Open the saved PNG (or describe it) to confirm all four curves and the capacity line are visible and legible before reporting completion.
\end{itemize}

\end{planbox}

\section*{SI Bundle S2: Structure of the \texttt{.epx} output bundle}
\label{si:bundle-structure}

Every \texttt{run}, \texttt{calibrate}, or \texttt{project} operation writes a self-contained \texttt{.epx} directory. Listing~\ref{lst:bundle-tree} shows the bundle produced for the no-vaccination baseline of the case study presented in the main text. The other three scenario bundles follow an identical layout.

\begin{lstlisting}[style=treestyle,label={lst:bundle-tree}]
no_vaccination.epx/
  manifest.json          # metadata, file schema, provenance
  config.yaml             # the configuration that produced this bundle
  compartments.parquet    # compartment counts per simulation, date, age group
  transitions.parquet     # daily transition counts per simulation, date, age group
  parameters.parquet      # per-simulation parameter values
  figures/
    hospital_capacity_comparison.png
\end{lstlisting}

\vspace{0.4cm}
\filename{manifest.json}. The manifest is a single JSON file recording everything needed to interpret the bundle without opening the Parquet files: the Epydemix version, model type and compartment list, the demographic groups used, the simulation window and number of stochastic realizations, the parameter values used to produce the run, a per-file schema (column names, dtypes, and array shape) for each Parquet file, usage hints for querying the bundle via the CLI or directly in Python, and a provenance block recording the path to the exact configuration file that produced the bundle.

For the baseline scenario shown here: \texttt{\seqsplit{epydemix\_version}} \texttt{1.3.1}; model \texttt{SEIR} with compartments \texttt{Susceptible, Exposed, Infected, Recovered, Vaccinated, Hospitalized}; 5 demographic groups (\texttt{0-4, 5-19, 20-49, 50-64, 65+}); simulation window \texttt{\seqsplit{2024-01-01}} to \texttt{\seqsplit{2025-05-01}} (487 daily timesteps), with $n=100$ stochastic simulations.

\filename{compartments.parquet}. One row per (simulation, date) pair --- $100 \times 487 = 48{,}700$ rows for this study --- indexed by \texttt{sim\_id} and \texttt{date}. For every compartment, the file stores one column per demographic group plus an aggregate \texttt{\_total} column, giving $6$ compartments $\times$ $6$ columns (5 age groups + total) $= 36$ data columns, plus the two index columns (38 columns total). Column naming follows the pattern \texttt{\seqsplit{<Compartment>\_<group>}}, e.g. \texttt{\seqsplit{Hospitalized\_65+}} or \texttt{\seqsplit{Hospitalized\_total}}.

\filename{transitions.parquet}. Same row structure and indexing as \filename{compartments.parquet}, but each column records the \emph{daily flow} between two compartments rather than a compartment's standing count, following the pattern \texttt{\seqsplit{<Source>\_to\_<Target>\_<group>}} (e.g. \texttt{\seqsplit{Infected\_to\_Hospitalized\_total}}, \texttt{\seqsplit{Vaccinated\_to\_Exposed\_total}} for vaccine-breakthrough infections). This is the file used to compute derived quantities not natively exposed by \texttt{inspect} or \texttt{compare}, such as the leaky-vaccine attack rate reported in Section~3.1 (``Case study: vaccination scenarios'') of the main text, which sums the \texttt{\seqsplit{Susceptible\_to\_Exposed}} and \texttt{\seqsplit{Vaccinated\_to\_Exposed}} flows over the simulation horizon.

\filename{parameters.parquet}. One row per simulation, recording the parameter values used in that stochastic realization. For a \texttt{run} with fixed parameters (as in this case study) the file carries only the \texttt{sim\_id} index, since every simulation shares the same parameter values already recorded once in \filename{manifest.json}; for a \texttt{calibrate} bundle, this file instead carries one column per calibrated parameter, recording the accepted posterior sample used in each simulation.

\filename{config.yaml}. A verbatim copy of the declarative configuration that produced the bundle, duplicating the \texttt{model}, \texttt{population}, \texttt{parameters}, \texttt{simulation}, and \texttt{initial\_conditions} blocks from the originating scenario file. Storing this copy inside the bundle, rather than only referencing the path in \filename{manifest.json}, makes the bundle self-contained: the exact configuration that produced a result travels with it even if the original configuration file is later modified or deleted.

\section*{SI Comparison S3: With-framework vs. without-framework}
\label{si:with-without-comparison}

This section reports run-by-run detail underlying the process and outcome comparisons in the main text, using the four with-framework runs and four without-framework runs described in the main text.

\subsection*{Process comparison}

Table~\ref{tab:si-process} reports mean $\pm$ std across the four runs of each condition, split by planning and execution phase.

\begin{table}[h]
\centering
\begin{tabular}{llcc}
\toprule
Phase & Metric & With framework & Without framework \\
\midrule
\multirow{2}{*}{Planning}
 & Duration (s) & $240.2 \pm 33.0$ & $504.0 \pm 168.0$ \\
 & Cost (\$)    & $1.43 \pm 0.17$  & $2.73 \pm 0.60$ \\
\midrule
\multirow{2}{*}{Execution}
 & Duration (s) & $342.2 \pm 59.0$ & $331.0 \pm 99.0$ \\
 & Cost (\$)    & $1.55 \pm 0.27$  & $2.16 \pm 0.86$ \\
\midrule
\multirow{2}{*}{Total}
 & Duration (s) & $582.5 \pm 82.6$ & $835.0 \pm 128.8$ \\
 & Cost (\$)    & $2.98 \pm 0.38$  & $4.89 \pm 0.78$ \\
\bottomrule
\end{tabular}
\caption{Planning, execution, and total cost and duration, with-framework vs. without-framework ($n=4$ per condition).}
\label{tab:si-process}
\end{table}

The without-framework condition is both slower and more expensive at every phase. The gap is proportionally larger in duration than in cost: planning takes roughly twice as long without the framework ($240$s vs. $504$s) but costs about $1.9\times$ as much ($\$1.43$ vs. $\$2.73$), while execution duration is nearly identical between conditions ($342$s vs. $331$s) despite execution cost still being $\sim\!40\%$ higher without the framework. This is consistent with the without-framework condition spending planning time reading Epydemix's source directly rather than a validated interface description, and then executing against a less certain plan.

\subsection*{Content comparison}

All eight runs converge on the same qualitative structure: no-vaccination and slow rollout both breach the 55{,}000-bed capacity line for roughly a month, rapid rollout and combined both clear it in every run, and rollout speed is the dominant lever when vaccination is the only intervention available.

\paragraph{Baseline.} No-vaccination peak occupancy is tightly clustered regardless of condition: $94{,}755$--$99{,}079$ across all eight runs (mean $97{,}441 \pm 2{,}202$).

\paragraph{Slow rollout.} Unlike rapid rollout, slow rollout shows no real cross-condition split (Table~\ref{tab:si-slow}): mean peak occupancy is within a few thousand beds between conditions ($81{,}277$ vs.\ $80{,}587$), and every one of the eight runs, in both conditions, still breaches the $55{,}000$-bed line for roughly a month. Slow rollout is the weakest lever tested and is never sufficient to clear capacity on its own, in either condition.

\begin{table}[h]
\centering
\begin{tabular}{lcc}
\toprule
Metric & With framework & Without framework \\
\midrule
Peak occupancy (mean $\pm$ std) & $81{,}277 \pm 4{,}279$ & $80{,}587 \pm 6{,}843$ \\
Peak occupancy (range)          & $78{,}223$--$87{,}318$ & $73{,}630$--$87{,}849$ \\
Attack rate \% (mean $\pm$ std) & $72.5 \pm 2.3$         & $71.6 \pm 4.1$ \\
Days over capacity              & 28--31 (all 4 runs)    & 26--31 (all 4 runs) \\
\bottomrule
\end{tabular}
\caption{Slow rollout outcomes, with-framework vs. without-framework.}
\label{tab:si-slow}
\end{table}

\paragraph{Rapid rollout.} Every run in both conditions clears capacity (0 days over), but with markedly different variance structure (Table~\ref{tab:si-rapid}). Without-framework runs cluster moderately around a higher mean; with-framework runs are mostly very low but include one outlier (a single run that assumed a visibly more conservative rate than its three siblings), which drives the wide with-framework standard deviation.

\begin{table}[h]
\centering
\begin{tabular}{lcc}
\toprule
Metric & With framework & Without framework \\
\midrule
Peak occupancy (mean $\pm$ std) & $12{,}634 \pm 21{,}707$ & $20{,}674 \pm 10{,}046$ \\
Peak occupancy (range)          & $1{,}090$--$45{,}161$   & $11{,}187$--$34{,}872$ \\
Attack rate \% (mean $\pm$ std) & $14.5 \pm 22.8$         & $25.9 \pm 10.6$ \\
Days over capacity              & 0 (all 4 runs)          & 0 (all 4 runs) \\
\bottomrule
\end{tabular}
\caption{Rapid rollout outcomes, with-framework vs. without-framework.}
\label{tab:si-rapid}
\end{table}

\paragraph{Combined strategy.} Combined outcomes are more consistent than rapid rollout in both conditions, and clear capacity in every run (Table~\ref{tab:si-combined}). With-framework runs reach a lower mean peak and attack rate than without-framework runs.

\begin{table}[h]
\centering
\begin{tabular}{lcc}
\toprule
Metric & With framework & Without framework \\
\midrule
Peak occupancy (mean $\pm$ std) & $19{,}944 \pm 7{,}428$ & $33{,}697 \pm 6{,}470$ \\
Attack rate \% (mean $\pm$ std) & $33.5 \pm 8.5$         & $42.2 \pm 6.4$ \\
Days over capacity              & 0 (all 4 runs)         & 0 (all 4 runs) \\
\bottomrule
\end{tabular}
\caption{Combined strategy outcomes, with-framework vs. without-framework.}
\label{tab:si-combined}
\end{table}

\paragraph{Assumed parameters.} The prompt specifies vaccination rates and NPI magnitudes only qualitatively (``low/moderate/high constant per-capita rate'', ``reduced school and workplace contacts for a few months''), so each run chose its own numeric values (Table~\ref{tab:si-params}). Rapid-rollout rate means are essentially identical between conditions ($1.025\%$ vs.\ $1.028\%$/day); the outlier in Table~\ref{tab:si-rapid} is a single with-framework run that picked $0.50\%$/day against a $1.00$--$1.30\%$ cluster from its three siblings, i.e.\ the same rate-ambiguity effect that in other batches has shown up on the without-framework side. Slow-rollout rate means are closer between conditions ($0.115\%$ vs.\ $0.181\%$/day) than the outcome table above might suggest, and NPI magnitudes are similar between conditions and less consequential for outcomes: the combined strategy clears capacity in all eight runs regardless of the specific school/work reduction or window length chosen.

\begin{table}[h]
\centering
\begin{tabular}{llccc}
\toprule
Parameter & Condition & Slow / school & Rapid / work & Combined / window \\
\midrule
Vaccination rate (\%/day) & With framework    & $0.115 \pm 0.015$ & $1.025 \pm 0.327$ & $0.425 \pm 0.075$ \\
Vaccination rate (\%/day) & Without framework & $0.181 \pm 0.064$ & $1.028 \pm 0.357$ & $0.464 \pm 0.097$ \\
\midrule
NPI reduction (\%)        & With framework    & $62.5 \pm 13.0$ & $47.5 \pm 10.9$ & $82.0 \pm 15.0$ days \\
NPI reduction (\%)        & Without framework & $55.0 \pm 5.0$  & $38.8 \pm 7.4$  & $85.0 \pm 8.7$ days \\
\bottomrule
\end{tabular}
\caption{Assumed vaccination rates (slow/rapid/combined) and NPI intervention magnitudes (school/work reduction, window length), mean $\pm$ std across runs.}
\label{tab:si-params}
\end{table}

\subsection*{Caveats and summary}

These are small samples ($n=4$ per condition), and a single differently-worded rate choice can flip a whole condition's headline outcome, as seen in the rapid-rollout outlier above. With-framework content metrics were transcribed by hand from each run's own result table, since the framework's bundle format has no single machine-readable summary file; without-framework metrics come directly from each run's results file.

The comparison does not change the picture in the main text: the without-framework condition remains costlier and more planning-heavy, baseline dynamics converge tightly regardless of condition, and unspecified vaccination-rate magnitudes remain the single largest source of cross-run disagreement on whether rapid rollout alone suffices. Both conditions reach the same conclusion on which strategies clear hospital capacity.

\section*{SI Guide S4: Step-by-step guide to reproduce the case study}
\label{si:guide}

This section walks through, command by command, how to reproduce the vaccination case study reported in the main text, starting from nothing but a terminal. No prior familiarity with Epydemix, the agent framework, or Claude Code is assumed.

\paragraph{What you will need.}
\begin{itemize}
  \item A computer with \texttt{Python} 3.9 or later and \texttt{git} installed.
  \item A terminal (Terminal.app on macOS, PowerShell/WSL on Windows, or any terminal on Linux).
  \item An AI coding agent capable of reading local files and issuing shell commands. This guide uses Claude Code, but the framework is agent-agnostic: any comparable terminal-based agent will work the same way, since it interacts with Epydemix purely through the CLI and \filename{AGENT.md}.
  \item An active subscription or API access for whichever agent you choose (e.g., a Claude account for Claude Code).
\end{itemize}

\paragraph{Step 1 --- Clone the repository and install the CLI.}
Open a terminal and run:

\begin{lstlisting}[style=bashstyle]
git clone https://github.com/epistorm/epydemix.git
cd epydemix
pip install -e ".[agent]"
\end{lstlisting}

We recommend doing this inside a fresh virtual environment (\texttt{python -m venv .venv \&\& source .venv/bin/activate} on macOS/Linux, or the \texttt{conda} equivalent) to avoid dependency conflicts with other Python projects on your machine.

\paragraph{Step 2 --- Verify the installation.}
Confirm the CLI is working before involving an agent:

\begin{lstlisting}[style=bashstyle]
epydemix models
\end{lstlisting}

This should print a JSON list of available model types (e.g. \texttt{SIR}, \texttt{SEIR}, \texttt{SIS}). If the \texttt{epydemix} command is not found even though installation succeeded, it may not be on your system \texttt{PATH}; in that case, replace \texttt{epydemix} with \texttt{python3 -m epydemix.cli.main} in every command in this guide --- the two forms are equivalent.

\paragraph{Step 3 --- Install and launch your AI agent.}
Install Claude Code following Anthropic's instructions for your platform, then launch it \emph{from the repository root} (the \texttt{epydemix} folder created in Step 1):

\begin{lstlisting}[style=bashstyle]
claude
\end{lstlisting}

Launching from the repository root matters: it is what allows the agent to discover \filename{AGENT.md}, the contract document that tells it how to use the framework. If you are using a different agent, the equivalent step is simply to start it with the repository root as its working directory.

\paragraph{Step 4 --- Give the agent the case-study prompt.}
Copy the full prompt exactly as given in Section~3.1 of the main text and paste it as your first message to the agent. Optionally, invoke the agent's planning mode first (in Claude Code, this is a mode toggle available before sending the message) to see its intended approach before it executes anything, as we did in SI Plan S1.

\paragraph{Step 5 --- Let the agent run.}
The agent will work autonomously through the discover $\rightarrow$ declare $\rightarrow$ run $\rightarrow$ inspect workflow: writing configuration files, validating them, launching simulations, and querying the results, without further input from you. This typically takes several minutes, depending on the agent and simulation settings. You do not need to do anything during this step, though you can watch the agent's terminal output to follow its progress.

\paragraph{Step 6 --- Locate the outputs.}
Once the agent finishes, it will report a summary directly in the chat and leave the full artifacts on disk under the working directory it created, typically:

\begin{lstlisting}[style=treestyle]
studies/italy_respiratory_virus/
  configs/     # the YAML configuration files the agent wrote
  bundles/     # one .epx output bundle per scenario (see SI Bundle S2)
  analyze.py   # the script that produced the comparison figure
\end{lstlisting}

The comparison figure itself sits inside the baseline bundle's \filename{figures/} subfolder, e.g. \filename{\seqsplit{bundles/no\_vaccination.epx/figures/hospital\_capacity\_comparison.png}}.

\paragraph{Step 7 (optional) --- Inspect the results yourself, without an agent.}
Everything the agent did is also directly usable by a human from the same terminal. For example, to see summary statistics for a single bundle:

\begin{lstlisting}[style=bashstyle]
epydemix inspect studies/italy_respiratory_virus/bundles/no_vaccination.epx summary -v Hospitalized_total
\end{lstlisting}

or to compare two bundles directly:

\begin{lstlisting}[style=bashstyle]
epydemix compare studies/italy_respiratory_virus/bundles/no_vaccination.epx \
  studies/italy_respiratory_virus/bundles/rapid_rollout.epx \
  -m peak,peak_date -v Hospitalized_total
\end{lstlisting}

This is the same command-line surface the agent used throughout; nothing about the bundles or the CLI is agent-specific, so any step in this guide can equally be carried out by hand.

\paragraph{Troubleshooting.}
\begin{itemize}
  \item \textbf{\texttt{epydemix: command not found}.} Use \texttt{python3 -m epydemix.cli.main} in place of \texttt{epydemix} throughout, as noted in Step 2.
  \item \textbf{The agent cannot find \filename{AGENT.md}.} Confirm you launched the agent from inside the cloned \texttt{epydemix} repository root, not a subdirectory or an unrelated folder.
  \item \textbf{Results differ slightly from those presented here.} Two independent sources of variation are expected. First, simulations are stochastic (Section~2, ``Materials and methods'', of the main text), so small numerical differences across runs of the same configuration are normal. Second, because the prompt specifies rollout speed and intervention magnitude qualitatively rather than numerically (Section~3.1 of the main text), a different agent run may choose somewhat different vaccination rates, NPI timing or combination, or initial conditions than those reported here, provided they remain internally consistent with the prompt's intent. The qualitative pattern --- rapid rollout and the combined strategy avoiding capacity breach, slow rollout and no vaccination both breaching it --- should reproduce reliably even so.
\end{itemize}

\section*{SI Additional Tasks S5: Experiments with Additional Tasks}
\label{si:additional-tasks}

\subsection*{Setup}
We compared two ways of driving Epydemix with a coding agent on a
small set of $5$ epidemiological modeling tasks. We compared a ``baseline'' condition, in which the agent only has access to the library, its documentation and its examples, against a ``framework'' condition in which the agent has access to our framework. We ran each of the $5$ tasks $5$ times for each condition ($2$ experimental conditions, $5$ tasks, $5$ independent iterations) for a total of $50$ agent sessions. Multiple iterations are necessary because the behavior of agents is stochastic.
The two conditions differ only in what we provide to the agent:
\begin{itemize}
\item \textbf{Baseline condition:} the Epydemix library from the GitHub repository with the agent framework removed (no \texttt{AGENT.md} contract, no command-line interface, and no references to them), together with its documentation and worked examples as Jupyter notebooks.
\item \textbf{Framework condition:} the Epydemix library as above, plus the agent framework: the \texttt{AGENT.md} contract and the framework tooling (command-line interface tool).
\end{itemize}

\subsection*{Agent harness}

Each task was executed by a non-interactive Claude Code session driven by a purpose-built agent harness and started by a single initial user prompt. No user interaction was possible. The settings of the session are reported in the Table~\ref{tab:si-harness} below:
\begin{table}[H]
  \centering
  \small
  \begin{tabular}{ll}
    \toprule
    Setting & Value \\
    \midrule
    Agent runtime      & Claude Code 2.1.224, non-interactive \\
    Model              & \texttt{claude-sonnet-5} \\
    Reasoning effort   & \texttt{medium} \\
    Tools available    & \texttt{Bash}, \texttt{Read}, \texttt{Write}, \texttt{Edit}, \\
                       & \texttt{Glob}, \texttt{Grep}, \texttt{BashOutput}, \texttt{KillShell} \\
    Tools withheld     & sub-agents, task planning, web search and fetch \\
    Local customization & disabled (\texttt{-{}-safe-mode}) \\
    Prompt             & single initial user message \\
    \bottomrule
  \end{tabular}
  \caption{Agent harness configuration.}
  \label{tab:si-harness}
\end{table}

The list of allowed tools was chosen so that both conditions have the capabilities required to carry out the tasks (running shell commands and writing/executing custom Python code) but nothing that would allow the agent to delegate or shortcut the work. Sub-agents and task-planning tools were withheld so that the measured effort is that of a single agent. Web access was withheld so the agent could not consult documentation outside of the Epydemix repository. The customization mechanism of Claude Code (e.g., skills, plugins, hooks, custom agents, external tool servers) were disabled using \texttt{-{}-safe-mode}, so that the local environment of the user could not influence the runs.

\subsection*{Isolation of task runs}

Several measures were taken to isolate the task runs from one another:

\begin{itemize}
\item \textbf{Workspace:} Each session received a fresh workspace from the Epydemix library's repository. Results, scripts and transcripts from earlier runs could not reach an agent's context.

\item \textbf{Environment:} Each task run received its own Python virtual environment, with the same set of data analysis packages together with the library's own dependencies (\texttt{numpy}, \texttt{pandas}, \texttt{scipy}). Plotting used a headless backend.

\item \textbf{Verification:} Task run isolation was verified. Every filesystem path appearing in tool call was checked, and any path outside of the execution workspace was flagged. Across the 50 sessions reported here, no session violated the isolation constraint.
\end{itemize}

\subsection*{Execution and metrics}

Tasks were run sequentially, with the two conditions (baseline vs framework) interleaved within each iteration, so that both met comparable constraints of Claude Code (API load, rate-limit conditions, etc.).
Metrics were taken from the structured event stream emitted by the agent runtime: wall-clock duration, time spent in model requests, the number of model turns, errors, and token usage decomposed into input, output, cache-read and cache-creation counts. Overall, the $50$ runs used 4.6 hours of agent wall-clock time and $\sim$84 million tokens.

\subsection*{Results: baseline vs framework}

The table below reports the median value of the chosen execution metrics for each of the $5$ tasks, for the baseline condition and the framework condition.

\begin{table}[H]
  \centering
  \small
  \begin{tabular}{ll rr}
    \toprule
    Task & Metric & Baseline & Framework \\
    \midrule
    SEIRHD scenarios & Wall clock time (s) & 160 {\scriptsize [146,\,177]} & 169 {\scriptsize [142,\,767]} \\
     & Time in API calls (s) & 146 {\scriptsize [126,\,153]} & 143 {\scriptsize [107,\,748]} \\
     & \# Turns & 30 {\scriptsize [24,\,37]} & 23 {\scriptsize [18,\,26]} \\
     & \# Output tokens & 12\,374 {\scriptsize [11\,015,\,14\,188]} & 9\,717 {\scriptsize [9\,094,\,10\,909]} \\
     & Cost (\$) & 0.785 {\scriptsize [0.698,\,0.865]} & 0.715 {\scriptsize [0.634,\,0.774]} \\
    \midrule
    SIR calibration & Wall clock time (s) & 214 {\scriptsize [163,\,237]} & 156 {\scriptsize [126,\,164]} \\
     & Time in API calls (s) & 146 {\scriptsize [126,\,158]} & 83 {\scriptsize [82,\,95]} \\
     & \# Turns & 39 {\scriptsize [37,\,44]} & 24 {\scriptsize [23,\,28]} \\
     & \# Output tokens & 12\,150 {\scriptsize [10\,383,\,13\,471]} & 7\,005 {\scriptsize [5\,988,\,7\,668]} \\
     & Cost (\$) & 0.892 {\scriptsize [0.673,\,0.981]} & 0.636 {\scriptsize [0.614,\,0.709]} \\
    \midrule
    Calibrate $\rightarrow$ Project & Wall clock time (s) & 226 {\scriptsize [215,\,263]} & 119 {\scriptsize [114,\,128]} \\
     & Time in API calls (s) & 197 {\scriptsize [193,\,232]} & 99 {\scriptsize [95,\,108]} \\
     & \# Turns & 43 {\scriptsize [40,\,49]} & 27 {\scriptsize [25,\,30]} \\
     & \# Output tokens & 17\,229 {\scriptsize [17\,162,\,20\,693]} & 7\,670 {\scriptsize [6\,527,\,7\,806]} \\
     & Cost (\$) & 1.308 {\scriptsize [0.929,\,1.384]} & 0.674 {\scriptsize [0.646,\,0.708]} \\
    \midrule
    School-closure sweep & Wall clock time (s) & 292 {\scriptsize [226,\,336]} & 594 {\scriptsize [377,\,860]} \\
     & Time in API calls (s) & 224 {\scriptsize [170,\,231]} & 314 {\scriptsize [281,\,425]} \\
     & \# Turns & 42 {\scriptsize [27,\,45]} & 58 {\scriptsize [42,\,73]} \\
     & \# Output tokens & 19\,970 {\scriptsize [11\,806,\,21\,624]} & 26\,312 {\scriptsize [23\,695,\,36\,137]} \\
     & Cost (\$) & 1.010 {\scriptsize [0.545,\,1.115]} & 1.824 {\scriptsize [1.358,\,2.519]} \\
    \midrule
    Measles coverage sweep & Wall clock time (s) & 644 {\scriptsize [544,\,1\,012]} & 519 {\scriptsize [346,\,668]} \\
     & Time in API calls (s) & 372 {\scriptsize [337,\,436]} & 308 {\scriptsize [243,\,344]} \\
     & \# Turns & 45 {\scriptsize [42,\,53]} & 36 {\scriptsize [33,\,40]} \\
     & \# Output tokens & 33\,065 {\scriptsize [30\,659,\,37\,857]} & 23\,486 {\scriptsize [21\,388,\,30\,821]} \\
     & Cost (\$) & 1.560 {\scriptsize [1.312,\,1.880]} & 1.371 {\scriptsize [1.195,\,1.487]} \\
    \bottomrule
  \end{tabular}
  \caption{Per-condition medians ($5$ runs per cell) with the observed minimum--maximum range in square brackets. \emph{Baseline} is the bare Epydemix library, and \emph{Framework} is the agent framework.}
  \label{tab:per-condition-medians}
\end{table}

% PREABLES + TASK PROMPTS
\input{task_prompts}

\end{document}

%% file: task_prompts.tex
\subsection*{Task prompts}

Each task prompt was assembled from three parts: a condition-specific orientation preamble, the task body, and shared closing instructions. The preamble is the only part of the prompt that differs between the two experimental arms. Task bodies are reported verbatim in the following section.

\begin{promptbox}[{{Preamble: Baseline condition (library only)}}]
\begin{Verbatim}[fontsize=\footnotesize, breaklines=true, breakanywhere=false, breaksymbolleft={}, breaksymbolindentleft=0pt, breakautoindent=false, breakindent=0pt]
You are working inside the epydemix repository. epydemix is a Python library for compartmental epidemic modeling. The library source is in `epydemix/`, the API documentation in `docs/`, worked examples as Jupyter notebooks in `tutorials/`, and an overview in `README.md`. The package is already installed in the active Python environment. Consult whatever you need, then carry out the following task.
\end{Verbatim}
\end{promptbox}

\begin{promptbox}[{{Preamble: Framework condition (agent framework)}}]
\begin{Verbatim}[fontsize=\footnotesize, breaklines=true, breakanywhere=false, breaksymbolleft={}, breaksymbolindentleft=0pt, breakautoindent=false, breakindent=0pt]
Read AGENT.md at the repo root, then carry out the following task.
\end{Verbatim}
\end{promptbox}

\begin{promptbox}[{{Closing instruction (identical in both conditions)}}]
\begin{Verbatim}[fontsize=\footnotesize, breaklines=true, breakanywhere=false, breaksymbolleft={}, breaksymbolindentleft=0pt, breakautoindent=false, breakindent=0pt]
Work only inside your current working directory -- that is the repository root for this session. Do not read or write anything outside it. Put every file you produce in a subdirectory called `outputs/` (create it). When you are done, give a short summary of your results in your final message.
\end{Verbatim}
\end{promptbox}

\subsection*{Task bodies}

\begin{promptbox}[{{T1 (SEIRHD scenarios): Custom SEIRHD model, three intervention scenarios}}]
\begin{Verbatim}[fontsize=\footnotesize, breaklines=true, breakanywhere=false, breaksymbolleft={}, breaksymbolindentleft=0pt, breakautoindent=false, breakindent=0pt]
MODEL:
Build a custom SEIRHD compartmental model for a respiratory pathogen in a population of 100,000:

S (Susceptible) -> E (Exposed) -> I (Infectious) -> H (Hospitalized) or R (Recovered) -> D (Dead, from H) or R (Recovered, from H)

EPIDEMIOLOGY:

- Transmission through contact between S and I (community), and at a reduced rate between S and H (nosocomial).
- Incubation period ~3 days (sigma = 0.33/day).
- Infectious period ~7 days (gamma = 0.14/day). At the end of it, 5% of cases are hospitalized, the rest recover.
- Hospital stay ~10 days (gamma_hosp = 0.1/day). Among hospitalized, 15% die, the rest recover.
- Community transmission rate: beta = 0.45/day. Nosocomial transmission rate: beta_hosp = 0.05/day.

SCENARIOS:
Run three scenarios over 6 months (Sep 1, 2024 -> Mar 1, 2025), each with 200 simulations.
Seed the outbreak with a tiny fraction of exposed and infectious individuals.

- Baseline -- no intervention.
- Early intervention -- on October 1, 2024, transmission drops 40% (beta goes from 0.45 to 0.27) until February 1, 2025.
- Late intervention -- same 40% reduction, but starting November 1, 2024 until February 1, 2025.

DELIVERABLES:

- Check that all three model setups are correct before running them.
- Run all three scenarios.
- Report, for each scenario, summary statistics and peak timing for I_total, H_total, D_total.
- Compute for each scenario: attack rate (%), peak hospital census (median + 90% CI), total deaths (median + 90% CI),
  days the hospital census exceeds 500 beds, lives saved vs. baseline.
- Produce three figures: epidemic curves (I_total), hospital capacity (H_total with a 500-bed line),
  cumulative deaths (D_total) -- all as scenario overlays with uncertainty bands.
- Summarize: what does a 30-day delay in intervention cost?
\end{Verbatim}
\end{promptbox}

\begin{promptbox}[{{T2 (SIR calibration): SIR calibration, posterior quality and two-parameter sensitivity}}]
\begin{Verbatim}[fontsize=\footnotesize, breaklines=true, breakanywhere=false, breaksymbolleft={}, breaksymbolindentleft=0pt, breakautoindent=false, breakindent=0pt]
STEP 1: Generate synthetic observed data

Run a forward SIR simulation with these "true" parameters to produce synthetic outbreak data:

- Population: 100,000 (default flat population)
- transmission_rate: 0.35
- recovery_rate: 0.1
- Period: Jan 1 - Apr 30, 2024 (120 days)
- Initial conditions: S = 0.999, I = 0.001, R = 0.0
- 1 simulation (or take the median of a few)

Extract the daily Infected_total time-series from this run and save it as a CSV file (observed.csv with
columns `date` and `cases`). This is the "observed data" we will calibrate against.

STEP 2: Calibrate transmission_rate

Set up a calibration that:

- Uses the same SIR model structure and recovery_rate = 0.1 (fixed).
- Defines a uniform prior on transmission_rate over [0.1, 0.8] -- deliberately wide, to test whether the
  calibration can recover the true value.
- Uses observed.csv as the observed data.
- Targets Infected_total as the comparison variable.
- Uses RMSE as the distance function.
- Uses the SMC strategy with 300 particles and 8 generations.

Define the base model setup once and reuse it for the calibration, so that the calibration only specifies
what it adds on top. Check the calibration setup is correct, then run it.

STEP 3: Assess calibration quality

- Inspect the posterior: what is the estimated transmission_rate? Report mean, median, std, and 95% CI.
  How close is the posterior median to the true value of 0.35?
- Inspect the calibration fit: get the fit trajectories (quantiles 0.05, 0.5, 0.95 for Infected_total) and
  compare them to the observed data.
- Produce a figure with two panels:
  - Left panel: posterior distribution of transmission_rate as a histogram, with a vertical line at the
    true value (0.35).
  - Right panel: calibration fit -- observed data as black dots, median fit as a blue line, 90% CI as a
    shaded band.

STEP 4: Sensitivity -- calibrate both parameters

Now set up a second calibration (reusing the same base model setup) that calibrates both
transmission_rate and recovery_rate:

- transmission_rate: uniform prior [0.1, 0.8]
- recovery_rate: uniform prior [0.03, 0.3]

Run this calibration (300 particles, 8 generations). Then:

- Inspect the joint posterior. Report the estimated values and compare to the true values
  (transmission_rate = 0.35, recovery_rate = 0.1).
- Produce a figure showing the joint posterior as a 2D scatter plot (transmission_rate on the x-axis,
  recovery_rate on the y-axis), with the true values marked as a red cross.

DELIVERABLES

- The observed.csv file with the synthetic data.
- Two validated calibration setups (single-parameter and two-parameter), both built on a shared base.
- Two sets of calibration results, with posterior and fit inspection reported.
- Both figures.
- A brief summary: did the calibration recover the true parameters? How tight are the posteriors? Did
  calibrating two parameters simultaneously make the estimates worse?
\end{Verbatim}
\end{promptbox}

\begin{promptbox}[{{T3 (Calibrate $\rightarrow$ Project): Calibrate, then project baseline vs intervention}}]
\begin{Verbatim}[fontsize=\footnotesize, breaklines=true, breakanywhere=false, breaksymbolleft={}, breaksymbolindentleft=0pt, breakautoindent=false, breakindent=0pt]
STEP 1: Generate synthetic observed data and calibrate

Simulate a synthetic outbreak (afterwards, we will calibrate against it).

Run a forward SIR simulation with these "true" parameters:
- Population: 500,000
- transmission_rate: 0.12
- recovery_rate: 0.04
- Period: Jan 1 - Mar 31, 2026 (~90 days)
- Initial conditions: S = 0.9999, I = 0.0001, R = 0.0
- 1 simulation

Extract Infected_total and save as observed.csv (columns: date, cases).

Run a calibration that:
- Fixes recovery_rate = 0.04.
- Places a uniform prior on transmission_rate over [0.03, 0.4].
- Uses the observed.csv generated above, targets Infected_total, uses RMSE distance.
- Uses the SMC strategy with 200 particles and 5 generations.

Validate and run the calibration. Inspect the posterior -- confirm that the estimated transmission_rate is
reasonably close to 0.12.

STEP 2: BASELINE projection (no intervention)

Produce a projection that extends the simulation 3 months beyond the calibration period:
- Starting from the simulation parameters above, change only the simulation end date to 30 June 2026.
- 200 posterior samples.

Inspect the results: what are the projected peak date and magnitude for Infected_total? Get quantiles
(0.05, 0.5, 0.95).

STEP 3: Intervention projection

Produce a second projection (intervention):
- Extend to 30 June 2026 (same as baseline).
- Override the transmission rate, dropping it to 0.03 from April 15 through June 1 (this should drive the
  effective R0 below 1, suppressing the epidemic two weeks into the projection window).
- 200 posterior samples.

STEP 4: Compare scenarios

Compare the projection results: baseline vs intervention. Report the comparison results and the deltas.

STEP 5: Visualize

Produce a single figure with two panels:
- Left panel: epidemic curves. Plot the median projected Infected_total for both scenarios (Baseline as
  blue, Intervention as orange) with 90% CIs as shaded bands. Mark the end of the calibration period
  (March 31) with a vertical dashed line. Mark the beginning and end of the intervention. If the observed
  data from Step 1 is available, overlay it as black dots for the calibration window.
- Right panel: cumulative attack rate. Plot the median cumulative Recovered_total (as a fraction of
  population) for both scenarios over the full projection period, with 90% CIs.

DELIVERABLES
- The synthetic observed.csv and a validated calibration output.
- Two projection outputs, each inspectable after the fact.
- The comparison of metrics (deltas) between the two scenarios.
- A two-panel figure comparing the scenarios.
- A brief summary: how much does the intervention reduce the peak? Delay it? Lower the final attack rate?
  Are the 90% CIs well-separated or do the scenarios overlap substantially?
\end{Verbatim}
\end{promptbox}

\begin{promptbox}[{{T4 (School-closure sweep): Optimal timing of a three-week school closure (NY State)}}]
\begin{Verbatim}[fontsize=\footnotesize, breaklines=true, breakanywhere=false, breaksymbolleft={}, breaksymbolindentleft=0pt, breakautoindent=false, breakindent=0pt]
Study the effect of a three-week school closure on a regular influenza season in New York State.

SETUP
- Use New York State's age-structured population and all available contact layers (home, school, work,
  community).
- Influenza-like epidemiology, seeded with a small number of infectious individuals at the start of the
  season.
- The season runs 1 October 2026 -> 31 May 2027.
- Tune the model so that, with no intervention, the epidemic peaks around 15 February 2027. Report the
  peak date you actually achieve and how you got there.
- Use 100 stochastic realizations per configuration.

INTERVENTION
A school closure removes all school-layer contacts for 21 consecutive days. Nothing else changes -- the
other contact layers are untouched.

QUESTION
Where in the season should those three weeks be placed to minimize the final attack rate, and how much
does the best placement actually buy you?

- Sweep the closure start date weekly from 1 December 2026 through 1 April 2027, running the full season
  for each start date.
- Also run the unmitigated season as the reference.
- For each start date compute the final attack rate (cumulative infections as a percentage of the
  population), the peak incidence, and the peak date.

DELIVERABLES
- A table of closure start date -> final attack rate, peak incidence, peak date.
- The optimal start date, and the relative reduction in attack rate it achieves versus no closure.
- How sensitive the benefit is to timing: how much of the benefit is lost by starting two weeks too early
  or two weeks too late, and the range of start dates that come within 10% of the optimum's benefit.
- Two figures:
  - Final attack rate vs. closure start date, with the unmitigated attack rate as a horizontal reference
    line and the unmitigated peak date marked.
  - Epidemic curves over the season for three cases overlaid -- no closure, optimally timed closure, and a
    closure starting one month after the optimum -- with the closure windows shaded and uncertainty bands.
- A short summary answering the question directly: what is the ideal placement, what relative impact does
  it achieve, and briefly why the optimum sits where it does.
\end{Verbatim}
\end{promptbox}

\begin{promptbox}[{{T5 (Measles coverage sweep): Measles risk in Vermont across vaccination coverage}}]
\begin{Verbatim}[fontsize=\footnotesize, breaklines=true, breakanywhere=false, breaksymbolleft={}, breaksymbolindentleft=0pt, breakautoindent=false, breakindent=0pt]
Study the potential impact of a measles outbreak in the state of Vermont if vaccination coverage falls
into the 80%-95% range.

SETUP
- Use Vermont's age-structured population and all available contact layers (home, school, work,
  community).
- Measles epidemiology in an SEIR structure: latent period ~12 days, infectious period ~8 days, and a
  transmission rate chosen to give a basic reproduction number typical of measles (R0 ~ 15). State the R0
  you actually obtain and how you determined it.
- MMR coverage is represented as immunity present before the outbreak starts: a covered individual is
  immune at day 0 and plays no further part in transmission.
- Seed the outbreak with 5 imported infectious individuals on 1 January 2027 and run through
  31 December 2027.
- Use 200 stochastic realizations per coverage level.

PART 1 -- Coverage sweep
Sweep coverage from 80% to 95% in steps of 1 percentage point. For each level report:
- the probability of a large outbreak (define a threshold, e.g. more than 1,000 cases, and say what you
  used),
- median total cases and the attack rate **among the initially susceptible** (do not count the
  pre-existing immune as cases),
- peak incidence and peak date, with uncertainty.

PART 2 -- Imperfect vaccine
The MMR vaccine is not perfect. Repeat the sweep assuming it is 97% effective, so 3% of vaccinated
individuals remain fully susceptible. Does any coverage level in the 80%-95% range still reach herd
immunity?

PART 3 -- Age structure and targeting
At 88% coverage (97% vaccine efficacy), report the age-stratified attack rate and the share of total cases
falling in each age group -- in particular the 0-4 group, who are largely too young to be fully vaccinated.

Then: if Vermont could run a catch-up campaign targeting a single age group, raising that one group's
coverage to 98% while the rest stay at 88%, which group should it target to minimize total cases? Compare
the candidate age groups, and say whether the best target is the group that carries the most cases or the
one that drives the most transmission.

DELIVERABLES
- A table of coverage -> outbreak probability, total cases, attack rate, peak incidence, peak date, for
  both the perfect and the 97%-effective vaccine.
- A table comparing the single-age-group catch-up campaigns.
- A figure: total cases and outbreak probability against coverage, for both vaccine assumptions, with the
  theoretical herd-immunity threshold (1 - 1/R0) marked as a vertical line.
- A figure: epidemic curves at three representative coverage levels spanning the transition, with
  uncertainty bands.
- A short summary: over what coverage range does outbreak risk collapse, how sharp is that transition, and
  how does the empirical threshold compare with 1 - 1/R0? What does an 80% versus a 95% Vermont look like
  in absolute numbers, and where should a catch-up campaign go?
\end{Verbatim}
\end{promptbox}